\documentclass[12pt]{article}

\usepackage{titletoc}
\usepackage{amsmath}
\usepackage{amssymb}
\usepackage{amsfonts}
\usepackage{amsthm}
\usepackage{mathtools}
\usepackage{mathrsfs}
\usepackage{bm}
\usepackage{bbm}
\usepackage{calc}
\usepackage{thmtools,booktabs}

\usepackage[table,xcdraw,dvipsnames]{xcolor}
\usepackage{graphicx}
\usepackage{psfrag}
\usepackage{epsf}
\usepackage{float}
\usepackage{multirow}
\usepackage{caption}
\usepackage{subcaption}
\usepackage{rotating}
\usepackage{tikz}
\usepackage{pgfplots}

\usetikzlibrary{patterns}
\pgfplotsset{width=10cm}

\usepackage[algo2e,ruled,vlined]{algorithm2e}
\usepackage{algorithm}
\usepackage[noend]{algpseudocode}

\makeatletter
\renewcommand{\ALG@name}{Algorithm}
\makeatother

\usepackage[T1]{fontenc}
\usepackage{textcomp}
\usepackage{lmodern}
\usepackage{enumerate}
\usepackage{url}
\usepackage{hyphenat}
\usepackage{lipsum}
\usepackage{natbib}

\usepackage[colorlinks,citecolor=blue,urlcolor=blue]{hyperref}
\usepackage[nameinlink]{cleveref}

\definecolor{pink}{rgb}{0.9,0.17,0.31}

\theoremstyle{plain}
\newtheorem{theorem}{Theorem}
\newtheorem{proposition}{Proposition}

\theoremstyle{definition}
\newtheorem{definition}{Definition}

\theoremstyle{remark}
\newtheorem{remark}{Remark}

\newcommand{\KL}{\mathrm{KL}}
\newcommand{\Unif}{\mathrm{Unif}}

\crefname{theorem}{Theorem}{Theorems}
\crefname{proposition}{Proposition}{Propositions}
\crefname{lemma}{Lemma}{Lemmas}
\crefname{corollary}{Corollary}{Corollaries}
\crefname{definition}{Definition}{Definitions}
\crefname{assumption}{Assumption}{Assumptions}
\crefname{example}{Example}{Examples}
\crefname{remark}{Remark}{Remarks}
\crefname{algorithm}{Algorithm}{Algorithms}
\crefname{section}{Section}{Sections}
\crefname{subsection}{Section}{Sections}

\numberwithin{equation}{section}
\allowdisplaybreaks[3]

\tikzset{
  declare function={
    gamma(\x)=sqrt(2*pi)*\x^(\x-0.5)*exp(-\x)*exp(1/(12*\x));
  }
}

\tikzset{
  declare function={
    tpdf(\x,\nu)=
    gamma(0.5*(\nu+1))/
    (sqrt(pi*\nu)*gamma(\nu/2))*
    (1+\x^2/\nu)^(-(\nu+1)/2);
  }
}

\tikzset{
  declare function={
    invgampdf(\x,\a,\b)=
    (\b/\x)^\a/\x/gamma(\a)*exp(-\b/\x);
  }
}

\tikzset{
  hatch distance/.store in=\hatchdistance,
  hatch distance=5pt,
  hatch thickness/.store in=\hatchthickness,
  hatch thickness=0.5pt
}

\def\argmax{\mathop{\arg\max}}

\newcommand{\mT}{\mathcal T}

\newcommand{\TT}{\mathbb{T}}

\newcommand{\Bel}{\mathrm{Bel}}
\newcommand{\Pl}{\mathrm{Pl}}

\newcommand{\E}{\mathbb E}

\newcommand{\R}{\mathbb R}

\defcitealias{asrm2024planned}{ASRM, 2024}
\defcitealias{asrm2020reserve}{ASRM, 2020}
\defcitealias{acog2014decline}{ACOG and ASRM, 2014}

\renewcommand{\d}{\mathrm d}
\renewcommand{\b}{\bm{\beta}}

\newcommand{\nhphantom}[1]{%
  \ifmmode
    \settowidth{\dimen0}{$#1$}%
  \else
    \settowidth{\dimen0}{#1}%
  \fi
  \hspace*{-\dimen0}%
}

\def\spacingset#1{%
  \renewcommand{\baselinestretch}{#1}%
  \small\normalsize
}

\spacingset{1.4}

\title{\Large Possibilistic Radial Transport for Approximate IM Inference}

\author{
  Jungeum Kim\thanks{
    Jungeum Kim is an assistant professor at North Carolina State University. Email: \texttt{jkim255@ncsu.edu}.
  } ~and~Percy S. Zhai\thanks{
    Percy S. Zhai is a PhD student at the University of Chicago Booth School of Business. \newline   The authors thank Ryan Martin and Sehwan Kim for helpful discussions. Any errors are the authors’
own.}}
\date{}

\begin{document}

\maketitle
\vspace{-1cm}
\begin{abstract}
\spacingset{1.05}
Probing the hypothesis space after seeing the data remains valid under possibilistic inferential models (IMs), provided the significance level stays fixed. The price is computation, as each plausibility is a supremum of the possibility contour over the hypothesis, and the contour itself is approximated at each queried parameter value. We propose a possibilistic radial transport, which hides the contour value of a parameter in the radius of its source point. When a transport that maximizes within-shell entropy is picked, sampling parameters covering a confidence cut becomes a matter of truncating the radius. We provide a deep learning algorithm that enforces the contour depth condition while maximizing the entropy within each shell. Our amortization makes coverage and power assessments of the learned approximation practical as well as predictive check of new datasets. We also use the sampler to construct a Bel--Pl spectrum for comparing and selecting interpretable hypotheses that satisfy a prescribed Bel--Pl decision criterion. In simulations the learned contours match or improve on ellipsoidal approximations to the cuts, while the coverage and power track the exact reference. Finally, we probe hypotheses about ovarian aging using synthetic AMH records, asking for each woman how many more years her median AMH level will remain above a specified reference value.
\end{abstract}


\section{Introduction}
After observing data, what can we say about uncertainty in an unknown parameter? Inferential models (IMs) unfold this story of uncertainty through belief and plausibility \citep{martin2015inferential_book}. Since the story is told after seeing the particular data in hand, it reads much like a Bayesian posterior, yet it requires no prior distribution. At the same time, its reliability is calibrated across repeated sampling, so that the resulting inference has frequentist validity. In this work, we focus on possibilistic IMs \citep{martin2026possibilistic}, which build belief and plausibility on a single, intuitive possibility contour.

The contour is defined by $\pi_z(\theta) = \mathsf{P}_\theta\{R(Z,\theta) \le R(z,\theta)\}$, the probability that a fresh dataset from $\theta$ ranks no higher than the observed one, where $R$ is a ranking function such as the relative likelihood. It is typically estimated by standard Monte Carlo (MC), sampling many $Z$ from $\mathsf{P}_\theta$. When the rank is costly to evaluate (for example, when it requires solving an MLE problem), so are the contour, plausibility, and belief. Our goal is to build a \emph{task-specific junior AI} that can be reused on new datasets to rapidly approximate the IM contour and the resulting hypothesis tests.

While not easy to compute, the belief and plausibility (Bel--Pl) in IMs have an important property that makes them worth the effort. Denote them by $\Pl_z$ and $\Bel_z$. A hypothesis $H$ is rejected when $\Pl_z(H)=\sup_{\theta\in H}\pi_z(\theta)$ is sufficiently small and accepted when $\Bel_z(H)=1-\Pl_z(H^c)$ is sufficiently large. This rule encodes that acceptance requires the corresponding $(1-\alpha)$ confidence region to lie inside $H$, while rejection requires it to be disjoint from $H$. The consequence is uniform validity \citep[Corollary~2]{cella2023possibility}; if we fix $\alpha$ in advance and use the same valid contour, validity is preserved as we combine or change hypotheses about the same parameter. That is, we can even adjust a hypothesis after seeing the data, or choose a modified hypothesis after seeing the result of a previous test!  

We find this inferential feature very appealing because it can offer \emph{statistical frustration prevention} for people who collect and analyze data. In a typical hypothesis test, we specify a hypothesis and $\alpha$ before observing the data, then perform the test once the data are available. Imagine spending time and money collecting data, only to find that the hypothesis does not receive a clear answer. Altering the data or the threshold to obtain a statistical rejection is common enough to have earned unflattering names such as p-hacking, N-hacking, and $\alpha$-hacking \citep{simmons2011false, reinagel2023hacking, koning2023post}. On the other hand, under the uniform validity described above, we can keep $\alpha$ fixed and continue asking different questions by changing the hypothesis. Then, fast contour computation would make this flexible hypothesis probing much easier by helping to check various hypotheses in a reasonable amount of time.

Our pursuit of fast IM inference builds on the work of \citet{martin2025efficient}, whose \emph{stitched contour} approximation uses parameters drawn from an \emph{inner credal distribution}. This replaces expensive data-space Monte Carlo, which requires normalization for each simulated dataset $Z$, with parameter-space sampling at the observed $z$. In its main implementation, however, the sampler is constructed by ellipsoidal approximations to contour level cuts, which we refer to as Martin's IMMC method. This approximation is in line with a possibilistic Bernstein--von Mises theorem \citep{martin2025asymptotic}. For small $n$, the $\alpha$-cuts need not be close to ellipsoids. Proposition~1 illustrates how this departure from ellipsoidal geometry can affect the resulting plausibility assessments in a fixed parameterization. The plausibility of a hypothesis that the original IM could reject can become arbitrarily close to $1$ under this ellipsoidal approximation, so a rejection that the original IM delivers is lost.

Here, we take a different route to pursue an efficient sampler that provides parameter samples for Martin's stitched contour estimation. Our junior AI is a  \emph{flexible} deep neural sampler that transports the core of the source distribution into a region with high contour values. We seek a sampler within a class of transports that send a source point $u$ to a parameter whose contour value is the source depth, $\pi_z(T_z(u))=1-\|u\|$. We establish the completeness of this class, that its pushforward distributions are exactly the inner credal distributions. With this ``radial transport'' formulation, the contour cuts now become an outcome of learning the sampler rather than a prerequisite for constructing it. Among the many ways of distributing probability within each contour shell, we choose the one that maximizes within-shell entropy, yielding the shell-uniform maximum-entropy inner credal (MIC) distribution. When the contour cuts are ellipsoidal and the same shell reference measures are used, this coincides with the inner probabilistic approximation used by Martin's IMMC.

Another central word of our work is ``amortized''. Rather than learning a sampler for a single fixed $z$, we seek an amortized transport that, once trained for a chosen likelihood model, can accommodate many datasets. We demonstrate this reuse in our ovarian aging application using synthetic AMH records, where the individuals are analyzed under a common likelihood model. The same trained sampler is reused across individuals to probe hypotheses about how long each person's median AMH level remains above a specified reference value. Amortization also enables validation through extensive simulation-based coverage and power assessments before the sampler is used in practice. When a new dataset arrives, Bayesian-like predictive checks can provide a preliminary screen before using the learned sampler for inference.

With fast hypothesis probing enabled, the next question is which of the interpretable hypotheses supported by the same contour one should report. As a first step, we draw a Bel--Pl spectrum that shows, at each contour depth, the fraction of the shell occupied by a hypothesis. The spectrum gives us a glimpse of the hypothesis geometry hidden behind the Bel--Pl endpoints, and indicating the Bel or Pl location on the same plot adds extra informativeness. We use this information to compare and select interpretable hypotheses that satisfy a prescribed Bel--Pl decision criterion. Among hypotheses meeting the acceptance criterion, we favor those with higher mean contour depth while limiting the gap between their inner credal probability and belief.

Deep neural networks have been used to learn samplers for statistical inference. A prominent application is Bayesian posterior sampling, where neural samplers offer an alternative to MCMC such as \citet{wang2026generative} and \citet{kim2025deep} to name a few. In generalized fiducial inference, \citet{li2020deep} proposed a fiducial autoencoder combined with a rejection step analogous to approximate Bayesian computation. To our knowledge, ours is the first deep neural sampler for IM inference, building on the connection between IM contour approximation and parameter-space sampling established by \citet{martin2025efficient}. On the inferential side, \citet{grunwald2024beyond} and \citet{koning2023post} use e-values to study data-dependent significance levels and post-hoc decision-making. Here, we keep $\alpha$ fixed and probe different hypotheses after observing the data.

In the current version of the manuscript, we assess the learned approximation empirically without establishing theoretical error bounds. 
Our experiments show comparable or improved contour accuracy relative to IMMC, rapid inference after training, coverage near the nominal level, and power and hypothesis selection consistent with the reference results. In our hypothesis-selection example, MIC mitigates belief underestimation and recovers the reference selection.

The paper is structured as follows. In Section \ref{sec:ranking_learning}, we briefly review possibilistic IMs to provide the context and a high-level description of our approach. The possibilistic radial transport is introduced in Section \ref{sec:phil}, together with our algorithm for training both non-amortized and amortized samplers. Section \ref{why_Q} discusses the Bel--Pl spectrum and hypothesis selection criteria. The numerical study in Section \ref{sec:ovarian} demonstrates the faithfulness and efficiency of our methods in an ovarian aging analysis. The paper concludes with Section \ref{sec:discuss}.

\section{Possibilistic Inferential Model}\label{sec:ranking_learning}
In 1930, Fisher introduced the fiducial argument with an intriguing ambition of obtaining a distribution expressing uncertainty about $\theta$ without specifying a prior \citep{fisher1930inverse}. Many statistical minds have since pursued this ambition in different forms, including Dempster--Shafer theory \citep{shafer2020mathematical}, generalized fiducial inference \citep{hannig2016generalized}, and inferential models (IM, \citealp{martin2010dempster}) to name a few. Here we focus on the possibilistic IMs, which integrate an imprecise probability over $\theta$ with frequentist validity \citep{martin2026possibilistic}. Its core inferential framework is Bel-Pl through the imprecise probability. 

The Bel--Pl representation used here has its roots in \citeauthor{dempster1967upper}'s \citeyearpar{dempster1967upper} construction of lower and upper probabilities through multivalued mappings, which \citet{shafer2020mathematical} developed into a general theory of evidence. This construction assigns probability to sets of possible parameter values, allowing uncertainty to be expressed without selecting a single probability distribution on the parameter space. When these random sets are nested, the resulting belief function is consonant, and its plausibility is a possibility measure determined by a contour. The normalized likelihood was studied as such a contour \citep{shafer1982belief,wasserman1990belief}, but this representation alone did not guarantee frequentist calibration. In possibilistic IMs, frequentist validity is obtained by
calibrating the (likelihood based) ranking function under the sampling model, which shares a similar idea with conformal prediction \citep{martin2026possibilistic}.

\subsection{The contour and Bel-Pl Inference}
Let $(\mathcal Z,\mathscr Z)$ be a measurable observation space, and let
$\{P_\theta:\theta\in\mathbb T\}$ be a statistical model, where each
$P_\theta$ is a probability measure on $(\mathcal Z,\mathscr Z)$. Let $P_\theta^{(n)}$ denote the distribution for $n$ i.i.d. samples $Z_n$ under
parameter $\theta$. 
The possibilistic IM contour is defined by 
\begin{equation}\label{eq:im_contour}
    \pi_{z_n}(\theta):=
P_\theta^{(n)}
\left(
R(Z_n,\theta)\le R(z_n,\theta)
\right),~\theta\in\mathbb{T},
\end{equation} 
where we consider the standard likelihood score $R(z_n,\theta) = {L_{z_n}(\theta)}/{\sup_{\nu}L_{z_n}(\nu)}$ as the rank. 
Intuitively, if $z_n$ were generated from $P_\theta^{(n)}$, its rank among samples from $P_\theta^{(n)}$ would be indistinguishable, and hence approximately uniform. In contrast, if $z_n$ is unlikely under $P_\theta^{(n)}$, $R(z_n,\theta)$ may become distinctively small. This uniformity gives the frequentist validity
\begin{equation}\label{eq:im_validity}
    \sup_{\theta\in \mathbb{T}} 
    P_{\theta}^{(n)} ( \pi_{Z_n}(\theta)\leq \alpha )
    \leq \alpha,
    \quad \text{for all } \alpha\in[0,1].
\end{equation}
Then we can construct $100(1-\alpha)\%$ confidence sets given by the $\alpha$-cuts of the contour defined as $C_\alpha(z_n)= \{\theta\in \mathbb{T}: \pi_{z_n}(\theta)\geq \alpha\}$. For hypothesis test with $\alpha\in[0,0.5]$, the rule is
\begin{equation}\label{eq:decision}
\begin{cases}
\operatorname{Bel}_z(H)\geq 1-\alpha, & \text{accept } H,\\
\operatorname{Pl}_z(H)\leq\alpha, & \text{reject } H,\\
\text{otherwise}, & \text{withhold judgment}.
\end{cases}
\end{equation}
Here, a low plausibility $\mathrm{Pl}_z(H):=\sup_{\theta\in H}\pi_z(\theta)$ indicates that all the points in \(H\) are not very compatible with the data, whereas a high belief, \(\mathrm{Bel}_z(H):=1-\sup_{\theta\in H^c}\pi_z(\theta)\), indicates that $H$ includes all the high compatibility parameter candidates. Importantly, because we must have $\max\{Pl_z(H), Pl_z(H^c)\}=1$, we always have either $Pl_z(H)=1$ or $Bel(H)=0$. Since then $Pl_z(H)\geq Bel(H)$, we know that $H$ cannot be accepted if $Pl_z(H)\neq 1$ and likewise $H$ cannot be rejected when $Bel(H)\neq 0$. 
\subsection{Stitched Contour Approximation}
The true contour $\pi_{z_n}(\theta)$ is difficult to evaluate, and a standard approximation is a Monte Carlo (MC) based approximation 
$\widehat{\pi}_{z_n}^{\mathrm{MC}}(\theta) = \frac{1}{M} \sum_{j=1}^M \mathbb{I}\bigl(R(Z_{n,j,\theta},\theta)\leq R(z_n,\theta)\bigr),$ for $\theta\in \mathbb{T}$ \citep{martin2015plausibility}. Then, constructing $C_\alpha$ is particularly expensive as it involves computing $\widehat\pi_{z_n}^{\mathrm{MC}}(\theta)$ for each  $\theta$ (on a grid) in $\mathbb{T}$, where for each $\theta$ extra sampling of $Z_{n,j,\theta}\sim P_\theta^{(n)}$ is conducted. Likewise, computing the supremum approximation for conducting the Pl/Bel test as in \eqref{eq:decision} is expensive.

Recent work of \cite{martin2025efficient} avoids repeating this data-space MC sampling by a parameter-space MC sampling reducing the computational cost significantly. Let $\mathcal P(\mathbb T)$ denote the collection of probability
measures on $(\mathbb T,\mathcal B(\mathbb T))$, and denote a possibility measure $\Pi_{z_n}(H)= \sup_{\theta\in H}\pi_{z_n}(\theta)$ (same as $Pl(H)$). He defines inner credal set as 
\[\mathcal C^{\mathrm{inn}}(\Pi_{z_n}):=\{Q\in\mathcal P (\mathbb{T}):Q(C_\alpha(z_n))=1-\alpha\text{ for all }\alpha\in[0,1]\}.\]
The definition implies that for $Q_{z_n}^{\mathrm{inn}}\in \mathcal C^{\mathrm{inn}}(\Pi_{z_n}),$ it is maximally consistent with the contour. Since  $(\pi_{z_n})_{\#}Q_{z_n}^{\mathrm{inn}} = \operatorname{Unif}(0,1)$, we emphasize that any $Q_{z_n}^{\mathrm{inn}}\in \mathcal C^{\mathrm{inn}}(\Pi_{z_n})$ has the calibration property 
\begin{equation}\label{eq:calibration}
  Q_{z_n}^{\mathrm{inn}}(\pi_{z_n}(\Theta)\le\alpha)=
\alpha,~\alpha\in[0,1].  
\end{equation}
Although the validity in \eqref{eq:im_validity} and the
calibration in \eqref{eq:calibration} involve
seemingly similar events, they refer to different sources of
randomness. The former is a sampling-calibration property under
$P_\theta^{(n)}$, with $\theta$ fixed and $Z_n$ random; the latter is
a depth-calibration property under $Q_{z_n}^{\mathrm{inn}}$, with $z_n$ fixed
and $\Theta$ random. Then his key innovation is to use this inner credal distribution as the sampler for the stitched contour estimate  \citep{cella2025computationally} as 
\begin{equation}\label{eq:Q_rho}
 \rho_{z_n,Q}(\theta)=
\mathbb{E}_{\widetilde\Theta\sim Q}
\left[\mathbb{I}\{r(z_n,\widetilde\Theta)
\le r(z_n,\theta)\}\right].
\end{equation}
with the parameter sampling distribution $Q$ as $Q^{\rm inn}$. Here, $r$ is unnormalized rank since the normalizing constant canceled out as this formulation considers only one $z_n$. Therefore, once having spent time to obtain $Q^{\rm inn}$, the empirical evaluation of \eqref{eq:Q_rho} for various $\theta$ in query is inexpensive, especially by re-using the samples from $Q^{\rm inn}$. 

Importantly, Corollary 1 of \cite{martin2025efficient} characterizes a $Q_{z_n}^{\mathrm{inn}}$ through a shell level Markov mixture $Q_{z_n}^{\mathrm{inn}} = \int_0^1 K^\alpha(\cdot)\d \alpha,$ where each $K^\alpha$ is supported on the boundary of $C_\alpha$. In this shell-mixture characterization, the $\alpha$-cut boundaries therefore serve as the starting point, naturally suggesting to obtain samplers by learning these cut boundaries. Approximating $Q_{z_n}^{\mathrm{inn}}$, \cite{martin2025efficient}
replaces the level set by an outer
ellipsoid for each $\alpha\in [0,1]$, and spread the weight by a uniform direction to the boundary. Note that such a $Q_{z_n}^{\mathrm{inn}}$ is not unique, and among them Martin seeks one that has a uniform distribution on each level-shell by a maximum entropy principle.

\subsection{Our Approach: Possibilistic Radial Transport}
The central theme of our work is viewing an inner credal distribution as a transport that preserves the contour depth. That is, when transporting the  source $u$, the depth constraint is $\pi_z(T_z(u))=1-\lVert u\rVert$. Then we render learning the maximum entropy inner credal distribution as finding a calibrated transport that maximizes the entropy among the transported things from the same depth of the source. This frees us from the need of estimating the boundary of the $\alpha$-cut of each level $\alpha\in[0,1]$ as our radial transport formulation does not  explicitly invoke level sets. In fact, the shell information has not disappeared; it simply has a \emph{hiding place} in the constraint linking source radius to contour depth. But then, any transport satisfying the depth constraint maps source samples with $\lVert U\rVert\le 1-\alpha$ into the corresponding $\alpha$-cut. That is, we use the transport to access the cuts rather than cuts as the pre-requisite for the credal distribution. 

\begin{figure}
    \centering
    \includegraphics[width=1\linewidth]{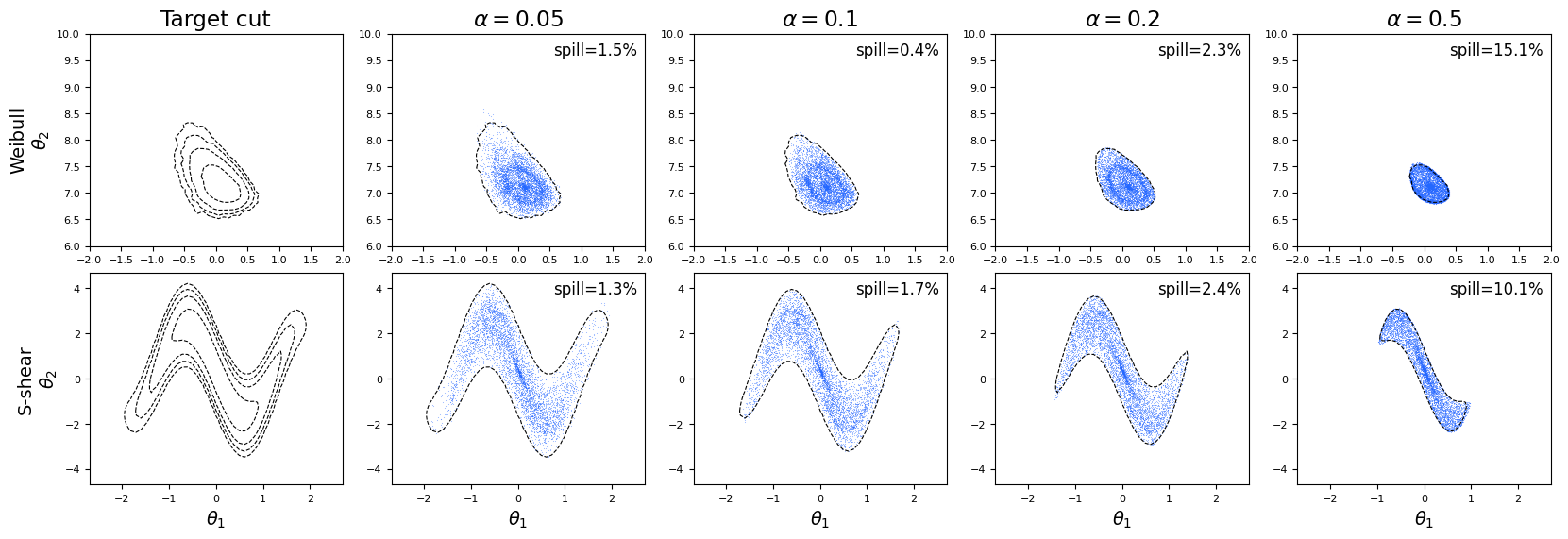}
   \caption{The nested structure of our trained radial transport $T^{\mathrm{MIC}}$. Transported samples from source balls of radius $1-\alpha$ approximately fill the corresponding $\alpha$-cuts. Black dotted lines indicate the estimated boundaries of the target cuts $C_\alpha$.}
    \label{fig:the_notion}
\end{figure}
When we seek maximum entropy principle, we maximize the conditional entropy of the transported source given the source radius, integrated over radii. Any maximizing transport pushes the source distribution to the maximum-entropy inner credal distribution, $Q_z^{\mathrm{MIC}}$. The hiding place, however, has its own computational challenge. Directly enforcing depth constraint during training would require expensive contour evaluations at each generated parameter value. We therefore replace the original contour with a differentiable stitched-contour approximation in the depth-matching loss. To keep this approximation faithful to the original contour, we add a faithfulness term that penalizes disagreement with contour targets precomputed at fixed training points. This allows us to train the transport without repeatedly evaluating the original contour. Figure \ref{fig:the_notion} shows how the trained transports map source balls of radius $1-\alpha$ to the corresponding $\alpha$-cuts. The transported samples closely follow the geometry of the target cuts, although learning error prevents an exact match.

To amortize the sampler w.r.t. $z$ as well, we need training target contour values for many datasets. Importantly, a likelihood normalizer depends on $Z$, not on the $\theta$ at which the likelihood is evaluated. Therefore, the training data for approximating $\pi_z(\cdot)$ for a given $z$ can be re-assembled to train an amortized sampler for approximating $\pi_Z(\cdot)$ for many different $Z$ (unknown at the time of training). Notice that Martin gives a direct sampler for nonelliptical star-convex cuts, which requires root-finding to locate a cut boundary for each sample \citep[Appendix~C]{martin2025efficient}. Compared to this version, then, the IMMC algorithm itself can be seen as an amortization, as approximating $\pi_z(\theta)$ at each $\theta$ does not require optimization personalized to $\theta$. In that sense our transport $T_z(u)$ for a fixed $z$ can be seen as an already amortized version w.r.t. $\theta$. In this regard, when we try to also amortize w.r.t. $z$, i.e., training $T(u,Z)$ for $Z$ unseen at the time of training, it is a \emph{double amortization}. 

\section{Possibilistic Radial Transport}\label{sec:phil}
In this section, we introduce the possibilistic radial transport, a map that hides the contour depth of a parameter in the radius of its source point. We establish that the pushforward distributions of this class are exactly the inner credal set $\mathcal{C}^{\mathrm{inn}}(\Pi_z)$, and that sampling from an $\alpha$-cut is recovered by simply truncating the source radius at $1-\alpha$. Among the transports in this class, we target the one that maximizes within-shell entropy. Then we develop a deep learning algorithm to find it and then amortize it across datasets.

\subsection{Depth-Preserving Radial Representation}\label{sec:radial_representation}
Our goal in this section is to build a flexible sampler for a distribution $Q\in\mathcal C^{\mathrm{inn}}(\Pi_z)$ that is uniform within each contour shell. We seek an $\alpha$-cut-free formulation of the sampler, which in turn lets us generate samples from the cuts.

We first define our target distribution. Consider the shell $L_{z,a} = \{\theta\in\TT:\pi_z(\theta)=a\}$, and consider the uniform distribution $\nu_{z,a} = \Unif(L_{z,a})$ over that shell.
Here, uniformity is defined relative to a declared shell reference measure. Assume that $a\mapsto\nu_{z,a}$ is a measurable probability kernel.
The maximum-entropy inner credal distribution is defined as
\begin{equation}\label{eq:def_MIC}
Q_z^{\mathrm{MIC}}(B)
=
\int_0^1 \nu_{z,a}(B)\,\d a.    
\end{equation}
Thus, \(Q_z^{\mathrm{MIC}}\) is a uniform mixture over contour depths and, conditionally on each depth, is uniform on the corresponding contour shell.
With the same choice of shell reference measures, this coincides with Martin's uniform-kernel inner probabilistic approximation.

The following proposition shows that replacing contour shells by the
boundaries of enclosing ellipses can substantially alter plausibility
assessments, even in the absence of numerical optimization and Monte Carlo
errors. Let $Q_z^E$ denote the population IMMC law obtained from the exact minimum-area MLE-centered enclosing ellipses, with uniform depths and uniform angular source directions. Recall the stitched contour $\rho_{z,Q}$ defined in \eqref{eq:Q_rho},
and write $\rho_z^{\mathrm{MIC}}=\rho_{z,Q_z^{\mathrm{MIC}}}$, $
\rho_z^E=\rho_{z,Q_z^E}.$ Also let \(\mathrm{Pl}_{\pi_z}(H)=\sup_{\theta\in H}\pi_z(\theta)\) be the plausibility under the original IM contour, and \(\mathrm{Pl}_{\rho_z^E}(H)=\sup_{\theta\in H}\rho_z^E(\theta)\) be the plausibility under the stitched contour computed from \(Q_z^E\).

\begin{proposition}[Distortion from elliptical approximation]
\label{prop:elliptical-distortion}
For every $\varepsilon\in(0,1/2)$, there exist a statistical model with parameter space $\mathbb R^2$, an observation $z$, and a half-space $H$
such that the exact IM contour $\pi_z$ has compact, connected cuts with smooth boundaries at every depth $a\in(0,1)$, for which $\rho_z^{\mathrm{MIC}}=\pi_z,$ $\operatorname{Pl}_{\pi_z}(H)\le\varepsilon$, and yet $\operatorname{Pl}_{\rho_z^E}(H)\ge1-\varepsilon$.
\end{proposition}
The proof is provided in Appendix~\ref{sec:proof-elliptical-distortion}.
A hypothesis rejected by the original IM can have plausibility arbitrarily close to one under the ellipsoidal approximation.
Thus, replacing the cuts by their enclosing ellipsoids may lead to an inferential conclusion different from that of the original IM, even when both the enclosing ellipses and the stitched contour are computed exactly. This motivates us to represent an inner
credal distribution without first approximating its shells by a prescribed family of shapes. 

We do this by tying $\mathcal C^{\mathrm{inn}}(\Pi_z)$ with a center-outward geometry by modeling through a transport from a \emph{source} noise distribution. Note that although $C_\alpha$ are nested by construction,  they do not necessarily have a center-outward geometry. Assume that $\TT$ is a Borel subset of $\R^d$ with $d\ge2$ and that $\pi_z$ is measurable.

\begin{definition}[Inner Radial Transportation]\label{def:radial}
Let $A\sim\Unif(0,1), V\sim\Unif(\mathbb S^{d-1}),
A\perp V,U=(1-A)V.$
Let $F_U$ denote the distribution of $U$.
Then $1-\lVert U\rVert=A\sim\Unif(0,1)$.
Define the class of inner radial transport as
\begin{equation}\label{eq:transport}
    \mT_{\mathrm{in}}(\pi_z)
=
\left\{
T:\mathbb B^d\to\TT:
T\text{ is measurable and }
\pi_z\{T(U)\}=1-\lVert U\rVert,~ F_U{-}a.s.
\right\}.
\end{equation}
\end{definition}
Interestingly, the pushforward distributions generated by $\mT_{\mathrm{in}}(\pi_z)$ form exactly the inner credal set, as the following theorem states. Furthermore, the cut-conditional distribution is recovered by transporting source draws conditional on $\lVert U\rVert\le1-\alpha$. This transport class also contains a map that generates the target $Q_z^{\rm MIC}$.

\begin{theorem}
\label{thm:main-radial}
The following properties of radial mapping holds:
\begin{enumerate}[(i)]
\item
The pushforward distributions generated by inner radial transports
form exactly the inner credal set $\left\{
T_\#F_U:
T\in\mT_{\mathrm{in}}(\pi_z)
\right\}
=
\mathcal C^{\mathrm{inn}}(\Pi_z).$
\item
For every $T\in\mT_{\mathrm{in}}(\pi_z)$ and $\alpha\in(0,1)$,
conditioning the source distribution on
$\lVert U\rVert\le1-\alpha$ yields the transported distribution
conditioned on $C_\alpha(z)$,
\[
T_\#
\left[
F_U\bigl(
\,\cdot\mid\lVert U\rVert\le1-\alpha
\bigr)
\right]
=
(T_\#F_U)\bigl(
\,\cdot\mid C_\alpha(z)
\bigr).
\]
In particular, there exists
$T^{\mathrm{MIC}}\in\mT_{\mathrm{in}}(\pi_z)$ such that $(T^{\mathrm{MIC}})_\#F_U=Q_z^{\mathrm{MIC}}.$
\end{enumerate}
\end{theorem}
The proof is provided in Appendix \ref{app:pf.main-radial}. As a characterization of $\mathcal C^{\mathrm{inn}}(\Pi_z)$, Theorem \ref{thm:main-radial} (i) is a transport counterpart to Martin's Corollary 1. We note that to find $Q$, we no longer have to explicitly approximate the cut boundaries $\partial C_\alpha(z)$ first. Rather than $\partial C_\alpha(z)$ appearing explicitly, the cuts have a “hiding place” through the depth constraint $\pi_z\{T(U)\}=1-\lVert U\rVert$. Thus, the radial representation has turned the task of directly constructing a shell mixture in $\mathcal C^{\mathrm{inn}}(\Pi_z)$ into one of finding a transport that obeys the depth constraint.

Now then, from this transport formulation, how can we pinpoint a $T^{\rm MIC}$ that pushes the source distribution forward to $Q^{\rm MIC}$? To this aim, we notice that the entropy maximization on each shell is possible while $\partial C_\alpha$ is still hiding, through the source radius condition. 
\begin{theorem}\label{thm:entropy_character}
Let $H_{z,a}$ denote entropy relative to the declared shell reference measure. We set entropy to $-\infty$ when absolute continuity fails.
Assume that this measure has finite, positive mass for almost every $a$. Also assume that
$\int_0^1
\left|H_{z,a}(\nu_{z,a})\right|
\,\d a
<
\infty.$
\begin{enumerate}[(i)]
\item For every $T\in\mT_{\mathrm{in}}(\pi_z)$, the following
entropy-gap identity holds:
\[\int_0^1
\left[ H_{z,a}(\nu_{z,a}) - H_{z,a} \left\{ T_\#\left[ F_U\bigl(\,\cdot\mid\lVert U\rVert=1-a\bigr) \right] \right\} \right]
\,\d a
=
\KL\left(T_\#F_U\Vert Q_z^{\mathrm{MIC}}\right).\]

\item For $T\in\mT_{\mathrm{in}}(\pi_z)$, the integrated conditional entropy is maximized if and only if the map sends the source distribution to $Q_z^{\rm MIC}$,
\begin{equation}\label{eq:TMIC}
T \in \argmax_{T\in\mT_{\mathrm{in}}(\pi_z)}
\int_0^1
H_{z,a}
\left\{
T_\#\left[
F_U\bigl(\,\cdot\mid\lVert U\rVert=1-a\bigr)
\right]
\right\}
\,\d a
\Longleftrightarrow T_\#F_U=Q_z^{\mathrm{MIC}}.
\end{equation}
\end{enumerate}
\end{theorem}
The proof is provided in Appendix~\ref{app:pf.entropy_character}. Theorem \ref{thm:entropy_character}(ii) follows from the completeness in Theorem~\ref{thm:main-radial}(i) and the entropy-gap identity in part (i), which shows that any departure from $Q_z^{\rm MIC}$ within the inner credal set strictly lowers the integrated conditional entropy. Note that the maximizing transport in \eqref{eq:TMIC} need not be unique (e.g., the source's rotation invariance). We take any maximizer and denote it as $T^{\rm MIC}$. The problem is now reshaped to finding a transport that maximizes the integrated conditional entropy among those obey the depth constraint.  

The hiding place for $\partial C_\alpha(z)$ is not cost free because enforcing $\pi_z\{T(U)\}=1-\lVert U\rVert$ is a serious headache in its own right. The depth constraint implies that we try to find $T^{\rm MIC}$, we have to evaluate $\pi_z$ after transporting every sample to make sure that a transport meets the restriction. Therefore, it is not that the challenge disappeared, but that it reshaped itself as another challenge. We think, however, that the challenge of evaluating $\pi_z$ is manageable by using a proxy to bypass these repeated contour evaluations, which does not require an ellipsoidal approximation of the cuts.

\begin{remark}[Kolmogorov$-$Smirnov diagnostics for depth calibration]\label{rm:remark_Dplus}
Let \(F_{\pi,Q}(\alpha)=Q\{\pi(\Theta)\le\alpha\}\), and define
$
D_+(Q)
=\sup_{\alpha\in(0,1)}[F_{\pi,Q}(\alpha)-\alpha]_+, $ and $D_-(Q)=\sup_{\alpha\in(0,1)}[\alpha-F_{\pi,Q}(\alpha)]_+.$ These are the one-sided Kolmogorov--Smirnov (KS) discrepancies between the distribution of $\pi(\Theta)$ under $Q$ and $\mathrm{Unif}(0,1)$. Replacing $F_{\pi,Q}$ with its empirical CDF yields the usual one-sample KS statistics. We report the two one-sided statistics separately to assess both the magnitude and direction of departures from uniform depth calibration. 
\end{remark}

\begin{remark}[Warning for a misconception]\label{rmk:warning}
We emphasize that $Q_z^{\rm MIC}$ is not defined by normalizing the contour to make it a distribution. For example, when $\pi_z$ is smooth with regular level sets, the coarea formula gives the density with respect to Lebesgue measure almost everywhere as $q_z^{\rm MIC}(\theta)=\|\nabla\pi_z(\theta)\|/|\partial C_{\pi_z(\theta)}(z)|.$\footnote{Here $|\partial C_a(z)|=\operatorname{Area}(L_{z,a})$, where $\operatorname{Area}$ denotes Euclidean
$(d-1)$-dimensional surface area in the chosen parameterization. Suppose $\pi_z$ is locally Lipschitz and
$0<|\partial C_a(z)|<\infty$ for almost every $a\in(0,1)$.
By the coarea formula,
$Q_z^{\mathrm{MIC}}(B)
=\int_0^1
\frac{\operatorname{Area}(B\cap L_{z,a})}
{|\partial C_a(z)|}\,da
=\int_B
\frac{\|\nabla\pi_z(\theta)\|}
{|\partial C_{\pi_z(\theta)}(z)|}\,d\theta,$ for every Borel set $B\subseteq\mathbb{T}$.} Intuitively, the pdf of $Q_z^{\rm MIC}$ would be higher where the contour slope relative to the level set size is larger and lower where this ratio is smaller. 
\end{remark}

\subsection{Radial Transport Map Learning}\label{sec:radial_learning}
In this section, we develop a learning algorithm for constructing an MIC sampler with target distribution $Q^{\mathrm{MIC}}_{z_n}.$ We now parameterize the transport as $T_\psi$, with $\psi\in\Psi$, and replace the exact depth constraint in \eqref{eq:transport} by a scoring-rule penalty. We compare two distributions through using a scalar scoring rule $S(p,q)$, where $p$ denotes the target value and
$q$ denotes the predicted value. This yields the following penalized population objective
\begin{equation}
\mathcal L(\psi)
\;=\;
-\int_0^1 H\bigl\{\,Q_\psi(\cdot\mid A=\alpha)\,\bigr\}\,d\alpha
\;+\;
\gamma_d\,\mathbb E\,\bigl[\,S\bigl(A,\;\pi_z\{T_\psi(U)\}\bigr)\bigr],
\label{eq:loss_Q}
\end{equation}for $U\sim F_U$, which is a sum of negative entropy with a depth-uniform penalty. Under exact feasibility, conditioning on $A=\alpha$ is equivalent to
conditioning on $\pi_z(T_\psi(U))=\alpha$. When feasibility is only approximate,
\eqref{eq:loss_Q} is a penalized relaxation rather than an equivalent
unconstrained formulation. In practice, we parametrize the transport $T_\psi$ by using a deep neural network.

At each iteration, we simulate a mini-batch of transported particles, and then perform stochastic optimization using empirical approximations to
\eqref{eq:loss_Q}. To address both differentiability and computational
efficiency, we will make extensive use of the (smoothed) stitched contour approximation. These approximations lead to the following empirical objective for stochastic optimization
\begin{equation*}
    \widehat{\mathcal L}(\psi)=
\gamma_h\widehat{\mathcal L}_h(\psi)
+\widehat{\mathcal L}_d(\psi)
+\gamma_f\,\widehat{\mathcal L}_{\mathrm f}(\psi),
\end{equation*}where the last term stems from replacing the true contour with its stitched approximation. Here we weight the entropy term by $\gamma_h$ rather than the depth term by $\gamma_d$ as in \eqref{eq:loss_Q}; the two are equivalent up to overall scale ($\gamma_h=1/\gamma_d$), and we found it more natural in practice to tune the entropy weight directly. In what follows, we will define and describe each term.

\paragraph{The first term: negative entropy.} The first term computes the negative entropy. Here, we adopt the Kozachenko--Leonenko entropy estimator
\citep{kozachenko1987sample,delattre2017kozachenko}. Given samples
$X_1,\dots,X_M$ from an $m$-dimensional distribution, it is defined by
\begin{equation}\label{eq:kl_entropy}
\widehat H(X_{1:M})=\frac{m}{M}\sum_{i=1}^M \log D^{(1)}_i+C_{m,M},
\end{equation}
where $D^{(1)}_i$ is the Euclidean distance from $X_i$ to its $1$-nearest neighbor among $\{X_j\}_{j\neq i}$, and $C_{m,M}$ is a bias-correcting constant depending only on $(m,M)$. In our setting, we compute
the entropy estimate among source points sharing the same radius, which has intrinsic dimension $m=d-1$. More precisely, let $A_i\stackrel{\rm iid}{\sim}
\operatorname{Unif}(0,1)$ and
$V_{j}\stackrel{\rm iid}{\sim}\operatorname{Unif}(\mathbb S^{d-1})$ for $i=1,...,N$ and $j=1,...,M$. Define $\widetilde{\Theta}_{ij,\psi}:=T_\psi((1-A_i)V_{j})$.
\[\widehat{\mathcal{L}}_h(\psi):=-
\frac1N\sum_{i=1}^N
\widehat H(\widetilde{\Theta}_{i1,\psi},\ldots,\widetilde{\Theta}_{iM,\psi}).
\]
Thus, $\widehat{\mathcal{L}}_h(\psi)$ includes both the within-shell entropy estimation and the Monte Carlo integration over the shell level.

\paragraph{The second term: uniform regularization.} The second term handles radial uniform regularization. For the scoring function $S$, we use the Brier-log composite score for $S$ as $S(p,q)
=
(p-q)^2
+\{-p\log q-(1-p)\log(1-q)\}.$ The Brier part stabilizes global shape matching, while the log-score part penalizes errors near 0 and 1, which are important for tail contours.
This score is minimized uniquely at \(q=p\). Note that any strictly proper scoring rule can
be used for the excess scores. 

Here, we must handle the issue that $\pi_z(\theta)$ may not be differentiable w.r.t $\theta$. For example, the simulator may not be written as a differentiable pathwise map w.r.t. $\theta$ such as discrete sampling or hard thresholding. The cost of computing $\pi_z(T_\psi(U))$ for each iteration is another challenge. Therefore, in the radial-uniformity term of \eqref{eq:loss_Q}, we replace
\(\pi\) by the stitched contour in \eqref{eq:Q_rho}. This is defined in our context as $\rho_\psi(\theta) := Q_\psi\!\left\{ R(\Theta)\leq R(\theta) \right\}.$ When estimating $\rho_\psi(\theta)$, we use differentiable, temperature-smoothed counterpart of the hard-indicator estimator, which is defined as 
\begin{equation}\label{eq:stitch_eg}
    \widehat\rho_{\psi,\tau}(\theta) = \frac1K \sum_{k=1}^K \sigma\!\left( \frac{ R(\theta)-R(\widetilde\Theta_k^{\mathrm{ref}}) }{\tau} \right), 
\end{equation}
where $\widetilde\Theta_k^{\mathrm{ref}} = T_\psi(U_k^{\mathrm{ref}})$, $U_k^{\mathrm{ref}}\overset{\mathrm{iid}}{\sim}F_U$,
\(\sigma(x)=(1+e^{-x})^{-1}\), and \(\tau>0\) is a temperature parameter. The
resulting empirical radial-uniformity loss is
\[ \widehat{\mathcal L}_d(\psi) = \frac{1}{NM} \sum_{i=1}^N \sum_{j=1}^M S\!\left( A_i,\, \widehat\rho_{\psi,\tau} (\widetilde\Theta_{ij}) \right).\]
The stitched computation requires no additional simulated data
\(Z\sim P_\theta\). Consequently, this replacement not only provides a
differentiable surrogate but also substantially reduces the computational cost
by avoiding repeated simulator draws and the associated normalization required
to evaluate the original contour for every generated query particle.

\paragraph{The third term: fidelity of the stitched proxy.} Since the depth constraint is enforced through the stitched approximation, it is very important that the stitched  proxy has a high fidelity to the contour. Conceptually, we want to penalize $\mathbb E_{\Theta\sim \operatorname{Unif}(\mathbb T)} \left[ S\!\left( \pi_z(\Theta),\rho_\psi(\Theta) \right) \right]$, speaking in the population level. Note that $\pi_z(\Theta)$ does not depend on $T_\psi$. Therefore, before
training, we construct the lookup table $\mathcal D_{\mathrm{target}} = \left\{ (\vartheta_r,\pi_z(\vartheta_r) ) \right\}_{r=1}^{N_{\mathrm{target}}},$ for $\vartheta_r\overset{\mathrm{iid}}{\sim} \operatorname{Unif}(\mathbb T).$ This one time construction helps to avoid repeatedly computing
these expensive contour values during training. At each iteration, we sample a mini-batch $\left\{ (\vartheta_\ell,\pi_z(\vartheta_\ell)) \right\}_{\ell=1}^{N_{\mathrm f}}$ from \(\mathcal D_{\mathrm{target}}\). As in the radial-uniformity loss, we use
the differentiable soft stitched estimate $\widehat\rho_{\psi,\tau}(\vartheta_\ell)$ as in \eqref{eq:stitch_eg}.
The resulting empirical fit loss is
\[ \widehat{\mathcal L}_{\mathrm f}(\psi) = \frac{1}{N_{\mathrm f}} \sum_{\ell=1}^{N_{\mathrm f}} S\!\left( \pi_z(\vartheta_\ell),\, \widehat\rho_{\psi,\tau}(\vartheta_\ell) \right). \]

After training, transporting source draws satisfying
$\|U\|\le 1-\alpha$ provides an approximate sampler for
the cut-conditional MIC distribution, without further
contour evaluations.  That is, while training $Q_\psi$ might be costly, once trained, (sub)sampling $C_\alpha$ is cheap.  The full algorithm for learning
\(Q^{\mathrm{MIC}}\) is presented in Algorithm~\ref{alg:mic} of
Section~\ref{sec:algs}.

\begin{remark}It is possible that the stitched contour approximates the original possibility contour very well. Suppose \(R_z(\Theta)\) has a continuous distribution under \(Q\).
Following \citet[Theorem~2]{martin2025no}
and using the probability integral transform,
\(\rho_Q=\pi_z\) if and only if
\(Q\in\mathcal C^{\rm inn}(\Pi_z)\) and
\(\pi_z=g_z\circ R_z\) for some nondecreasing \(g_z\). This establishes structural compatibility of the
objectives with an exact MIC transport that minimizing the transport loss while also minimizing the faithfulness loss at the same time is possible.
If \(\pi_z\) is not a nondecreasing function of \(R_z\),
no stitched proxy based on \(R_z\) can reproduce it exactly, and the
fidelity term may compete with the MIC objectives.
\end{remark}

\begin{figure}
    \centering
    \includegraphics[width=0.8\linewidth, height=7cm]{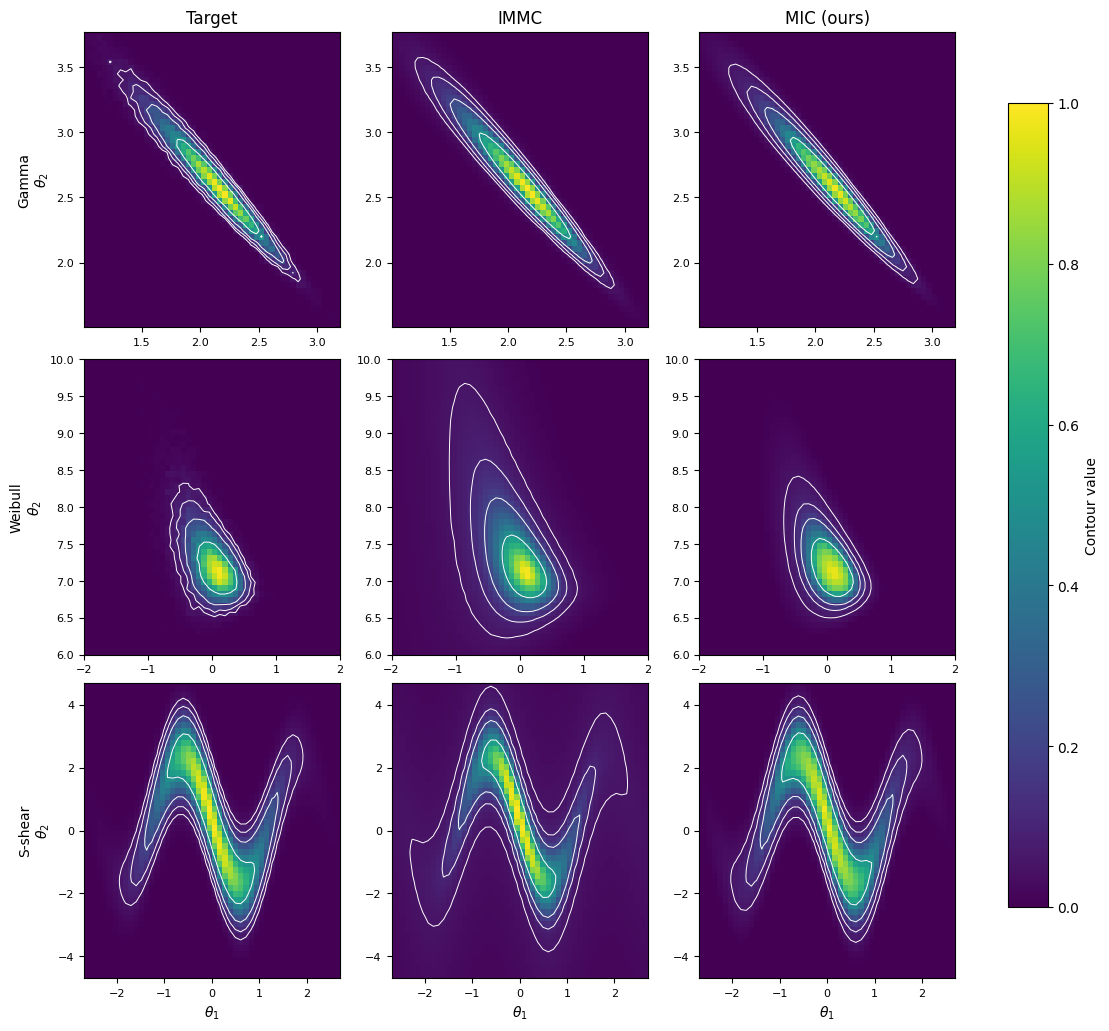}
    \caption{(Example~1) Target contour $\pi$ (left), IMMC estimate (middle) and MIC estimate (right) for Gamma, Weibull and S-shear. White curves are the level sets $\alpha\in\{0.05, 0.1, 0.2, 0.5\}$ of each panel.}
    \label{fig:three_toy_contour}
\end{figure}
\paragraph{Example 1.} Consider three toy examples of Gamma and Weibull \citep{martin2025efficient}, and Shear. For Gamma, we model the rat survival times as $Z_i \sim \mathrm{Gamma}(a,b)$ iid, with shape $a>0$, scale $b>0$, and log-parameter $\theta=(\log a,\log b)$ used in \cite{martin2025efficient}. For Weibull, we model the latent failure times as $Y_i \sim \mathrm{Weibull}(k,\lambda)$ iid, with shape $k>0$, scale $\lambda>0$, log-parameter $\theta=(\log k,\log \lambda)$, and observe right-censored data $(X_i,\Delta_i)$, where $X_i=\min(Y_i,C_i)$ and $\Delta_i=I(Y_i>C_i)$. As a non-star-convex demonstration, we include the Shear model whose contour is exactly known. Let $Z_i \overset{iid}{\sim} N_2(m(\theta), \sigma^2 I_2)$ for $i=1,...,10$ with $\sigma=2.5$ and $m(\theta) = (\theta_1,\, \theta_2 + 2\sin(2.5\theta_1))$. In $\eta = m(\theta)$ this is a Gaussian location model with closed-form contour $\pi_{z^n}(\theta) = \exp\{-n\|\bar z - m(\theta)\|^2/(2\sigma^2)\}$, whose $\alpha$-cuts are discs in $\eta$ and bent bands in $\theta$. We keep the focus on $\theta$, where the contour is non-star-convex.

\begin{table}
\centering
\caption{(Example~1) depth calibration, contour accuracy, and spatial coverage.}
\label{tab:example1_metrics}
\small
\setlength{\tabcolsep}{4pt}
\begin{tabular}{llrrrrrr}
\toprule
Model & Method
& \(D_+\) & \(D_-\) & MSE & max err. & Spearman & $C_{.05}$ cov. \\
\midrule
Gamma
& IMMC       & 0.070 & 0.026 & 0.0002 & 0.116 & 0.988 & 0.983 \\
& MIC (ours) & 0.075 & 0.063 & 0.0001 & 0.069 & 0.989 & 1.000 \\
\midrule
Weibull
& IMMC       & 0.153 & 0.025 & 0.0021 & 0.159 & 0.987 & 0.982 \\
& MIC (ours) & 0.136 & 0.056 & 0.0002 & 0.152 & 0.986 & 1.000 \\
\midrule
S-shear
& IMMC       & 0.034 & 0.136 & 0.0016 & 0.110 & 1.000 & 0.543 \\
& MIC (ours) & 0.057 & 0.049 & 0.0001 & 0.040 & 1.000 & 0.982 \\
\bottomrule
\end{tabular}
\end{table}

Figure \ref{fig:three_toy_contour} shows that MIC's reconstruction is visually comparable to that of IMMC for Gamma and more accurate for Weibull and S-shear, with tighter tail contours. Despite using ellipsoidal approximations, IMMC provides good reconstructions of non-elliptical contours away from the tails. The reconstructions are compared numerically in Table \ref{tab:example1_metrics}. For the uniform calibration check, we estimate $D_+$ and $D_-$ defined in Remark \ref{rm:remark_Dplus}. A conservative 5\% cutoff for the two-sided KS diagnostic is approximately $0.096$ (See, Section \ref{sec:details_example1} for details). We see that IMMC's $D_+$ and $D_-$ values are below this threshold for the Gamma model, indicating no detected departure from uniform calibration at this threshold. MIC's deviations fall below this threshold in Gamma and S-shear. For contour quality, we see that both methods perform reasonably well in terms of MSE, maximum absolute error, and Spearman correlation between the reconstructed contour $\hat{\rho}_Q$ and the target contour $\pi$. The coverage metric shows that MIC captures most of the grid-based approximation to $C_{.05}$. This is supported by the scatter plot of $Q$ samples in Figure \ref{fig:full_nested_contour} in the Appendix. Implementation and evaluation details for MIC and IMMC are provided in Section \ref{sec:details_example1}.

\subsection{Amortization w.r.t. Data $z$}

Here, we discuss how to learn a sampler $Q^{\rm MIC}_z(\theta)$ that is amortized not only w.r.t. $\theta$ but also w.r.t. the data $z$. That is, instead of learning a transport $T_z(u)$ for a fixed $z$, we can learn $T(u,z)$, where $z$ is also an input. Once trained, the conditional transport can be reused across datasets without training a separate sampler for each dataset, facilitating repeated-inference tasks such as power and coverage estimation besides the contour estimation and $\alpha$-cut sampling for the original $z_0$. It may bounce a new $z$ if its predictive distribution behaves differently from typical ones that are generated by the likelihood model we consider.

Amortizing our algorithm is not conceptually difficult. Once we have training dataset, the pair of $(z,\theta,\pi_z(\theta))$ for various $z$ and $\theta$ is given, our original learning algorithm extends to the conditional transport with a slight tweak to handle various $Z$ as described in Algorithm \ref{alg:mic_amortized}. Once the amortized sampler is trained, we can empirically assess the coverage of its confidence regions using independent simulated datasets before inserting this amortized machine into the real world. We can also estimate power to understand the sensitivity of the tests of interest under the specified likelihood model.  

While things sound simple, the cost of getting the training dataset is the overhead. Preparing the training data even for our standard MIC learning is computationally costly because it requires approximating the target $\pi_z(\theta)$ by MC. Therefore, thinking of obtaining the training data that will be used for even various $z$ realizations sounds daunting. Thankfully, the good news is that the training data preparation for the standard MIC learning \emph{already computes} the likelihood normalizing constants for any $Z$ that ever was sampled during the MC. For each simulated dataset $Z$, the likelihood normalizing constant depends only on $Z$, not on the parameter value $\theta$ at which the likelihood is evaluated. These resources can therefore be reused to construct training labels (contour values estimated through MC) across these datasets without additional simulation or MLE fitting. 
This reuse can substantially reduce the cost of generating the amortized training data, especially when likelihood normalization is the dominant cost. Utilizing this, an efficient training-data generation process is described in Algorithm \ref{alg:mic_amortized_data}.

A caveat is that the parameter box is chosen using an initial dataset $z_0$, so its suitability for amortized inference depends on whether it also covers the relevant contour regions for other datasets. When these regions are largely shared across datasets, as in the S-shear example, the same box can support amortized learning. However, when the feasible parameter region depends on $z$, the coverage of the box and the handling of support constraints would require additional care. Under a fixed computational budget, a wider box may accommodate more datasets but provide fewer training points with non-negligible contour values for each dataset. A narrower box concentrates training effort near the relevant parameter region for $z_0$, at the cost of a more limited range of reuse.

When we do not want to extend our training parameter space too much, we can resort to an internal predictive check we explain below. This can be a useful optional screening before applying the learned sampler to a new dataset. A discrepancy can reflect limitations of the chosen
parameter box, model misspecification, or inadequate generalization of the learned transport. While the check does not
distinguish these causes, it flags cases requiring further
investigation before interpreting the sampler’s output.

Our internal predictive check is 
a double generational predictive comparison to bounce a new dataset whose relationship with its predictive samples is atypical under our trained sampler.  We simulate the first generation (children) similar to Bayesian predictive sampling, i.e.\ sample $\theta_i\sim Q(\theta\mid z_0)$ and then sample $Z_i\sim P_{\theta_i}$. Then we generate the next generation (grandchildren) $Z_{ij}\sim P_{\theta_{ij}}$ for $\theta_{ij}\sim Q(\theta\mid Z_i)$, $j=1,\dots,J$. Here, the premise is that, if our sampler is a perfect sampler and if $z_0$ is from our specified likelihood model, the relationship between $z_0$ and its children should be similar to the relationship between each $Z_i$ and its children. This is simply because we know that $Z_i$ genuinely come from the model. This gives us a basic screening check. We measure this generational relationship by the distance between $Z_i$ and $Z_{ij}$, defining $d_{ij}=d(Z_i,Z_{ij})$. Here, since $Z$ itself is a set of observations, we use the energy distance \citep{szekely2013energy}, which coincides with the squared MMD with kernel $k(x,y)=(\|x\|+\|y\|-\|x-y\|)/2$ \citep{sejdinovic2013equivalence}. It is widely used to test whether two samples come from the same distribution, and is zero if and only if the two distributions coincide. Then, for each $i$, we can make a KDE plot using $\{d_{ij}\}_{j=1}^J$. We can compare these KDE curves with the KDE curve of $d_{0i}=d(z_0,Z_i)$, $i=1,\dots,I$. For example, we define a diagnostic score as the fraction of reference medians that are at least as large as the median distance from the observed data to its predictive samples.  We use this comparison as a heuristic diagnostic.

\paragraph{Example 2.} In demonstrating amortized learning, we focus on S-shear models. This is because the true contour is available in closed form, allowing direct evaluation (not training) of the learned contour and transport. It also provides a challenging amortization problem because the target contour geometry varies substantially with the observed data $z$. In particular, its central ridge can tilt in opposite diagonal directions for different datasets. We use a single trained conditional transport $T(u,z)$ throughout this
experiment, while IMMC is fitted
separately to each dataset. We center the transport at the MLE and condition its neural component on $(\sin(b\bar z_1), \cos(b\bar z_1))$, where $b=2.5$ is the known shear frequency and $\bar z_1$ is the sample mean of the first coordinate. This model-specific phase encoding reflects the changing shear geometry and allows us to retain a small network comparable to the non-amortized version. The implementation details are in Section \ref{sec:exp2_details}.
\begin{figure}
    \centering \includegraphics[width=0.75\linewidth]{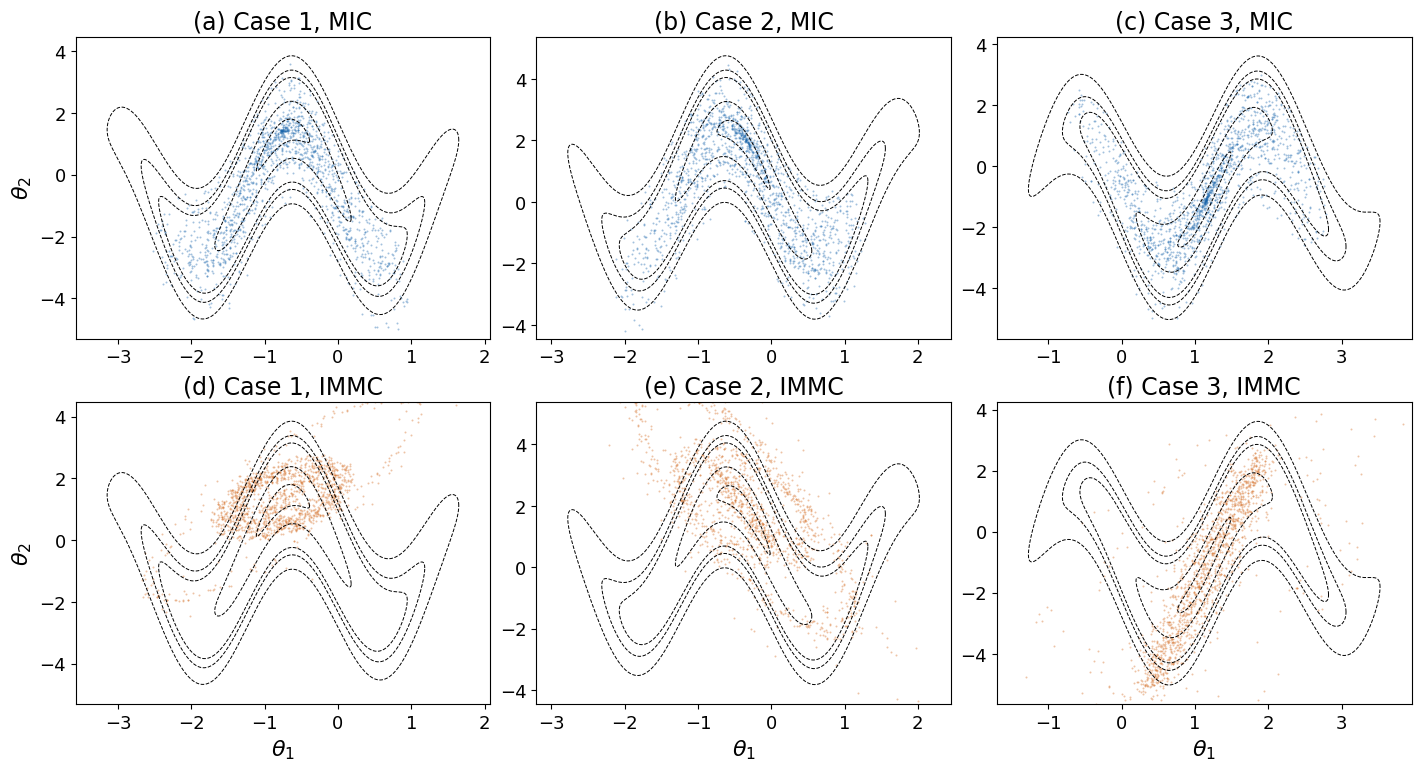}
    \includegraphics[width=0.24\linewidth]{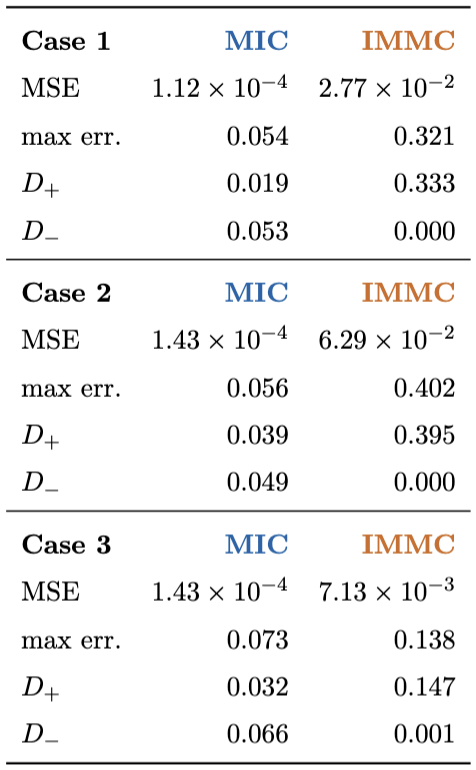}
    \caption{Contours of $\pi_z$ for three data realizations $z$
(cases 1--3). (a--c) Samples from a single amortized MIC sampler.
(d--f) IMMC samples, with a separate IMMC approximation fitted for
each dataset. (g) Contour approximation errors 
and depth-calibration statistics.}
    \label{fig:shear_amortize}
\end{figure} 

\textbf{1. Contour estimation.} Figure~\ref{fig:shear_amortize} compares the two samplers for three
data realizations. The MIC samples follow the curved target contours
despite their changing orientation and shape across datasets. The substantially smaller MSE values for MIC in Figure~\ref{fig:shear_amortize}g implies that a single transport map $T(u,z)$ maintains substantially smaller
contour errors across these examples. 
Figure~\ref{fig:sub1} in Appendix further shows that
MIC is closer to the desired depth calibration.

\begin{figure}
    \centering
\includegraphics[width=\linewidth, height=4.5cm]{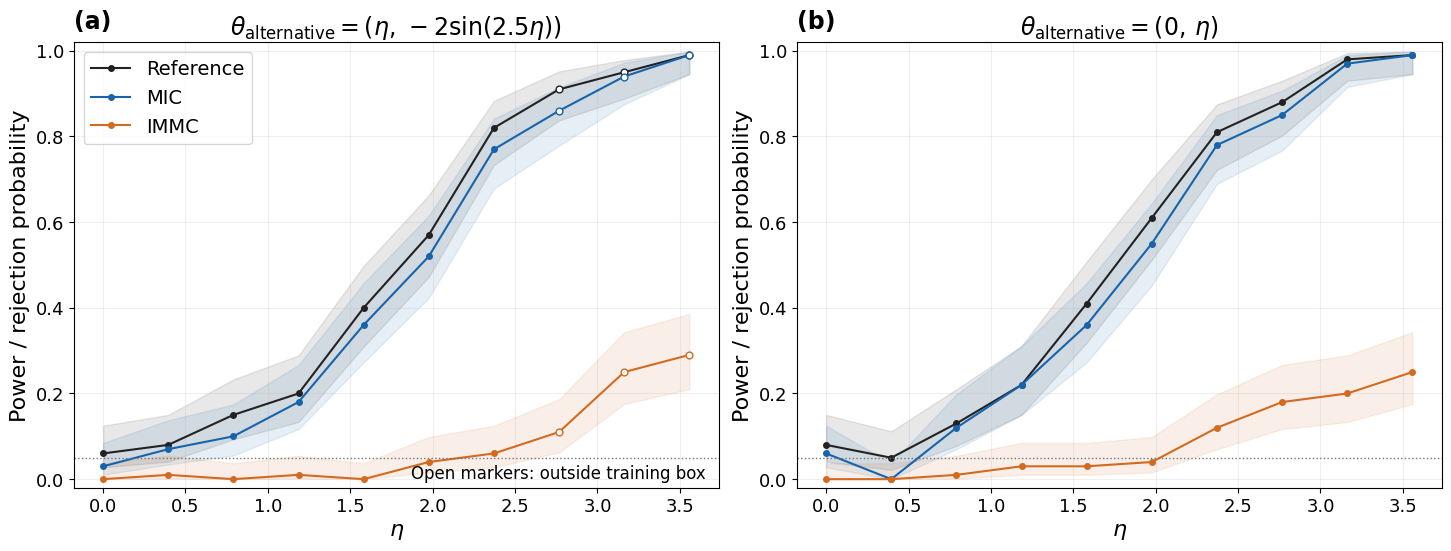}
    \caption{Empirical power for testing \(H_0:\theta=(0,0)\) in the S-shear model at \(\alpha=0.05\). We generate data from \(\theta_\eta=(\eta,-2\sin(2.5\eta))\) in the left panel
and \(\theta_\eta=(0,\eta)\) in the right panel, 100 times for each $\eta$. The empirical
type-I error rate corresponds to \(\eta=0\).}
    \label{fig:amortized_samples}
\end{figure}

\textbf{2. Coverage and Power estimation.} Table~\ref{tab:amortized_cost} summarizes coverage and efficiency
at true parameters $\theta_\phi=\bigl(\phi/2.5,-2\sin\phi\bigr)$
for $\phi\in\{0,\pi/2,\pi\}$.
For each fixed $\phi$, we generate 200 independent datasets and
evaluate coverage and the area of the 95\% confidence region.
At the nominal 95\% level, MIC has empirical coverage of 96.5--98.0\%.
MIC therefore exhibits mild empirical conservatism at this level,
with 95\% confidence-region areas close to those of the exact reference.  On the other hand, Figure~\ref{fig:amortized_samples} reports empirical rejection
probabilities for testing $H_0:\theta=(0,0)$ at level $\alpha=0.05$
under two families of alternatives,
$\theta_\eta^{(1)}=\bigl(\eta,-2\sin(2.5\eta)\bigr)$ and
$\theta_\eta^{(2)}=(0,\eta)$, which give
$m(\theta_\eta^{(1)})=(\eta,0)$ and
$m(\theta_\eta^{(2)})=(0,\eta)$, respectively.  Posing an alternative hypothesis $\theta_a$ as the ground truth, we can simulate data from $P_{\theta_a}$ many times and check how many times the hypothesis in test is rejected. Each rejection probability is estimated from 100 independent datasets.
We can see that MIC closely tracks the rejection probabilities of the exact reference. Interestingly, MIC remains close to the reference even when the true parameter values lie outside the
training box (those marked by open symbols). While this empirical generalization beyond the training parameter box
is interesting, we do not establish a guarantee for
such extrapolation.  

\begin{table}
\centering
\caption{(Amortized S-shear) Repeated inference across datasets, Coverage entries are estimates [pointwise 95\% Wilson CI]; area entries are medians [10th, 90th percentiles] across datasets. Reference is computed by using the exact, closed form.}
\label{tab:amortized_cost}
\footnotesize
\setlength{\tabcolsep}{3pt}
\begin{tabular}{llcccc}
\toprule
Phase & Method & 90\% coverage & 95\% coverage & 99\% coverage & 95\% area \\
\midrule
\(0\) & Reference & 0.920 [0.874, 0.950] & 0.955 [0.917, 0.976] & 0.995 [0.972, 0.999] & 11.764 [11.764, 11.764] \\
 & MIC & 0.945 [0.904, 0.969] & 0.980 [0.950, 0.992] & 0.995 [0.972, 0.999] & 13.103 [12.410, 13.552] \\
 & IMMC & 0.990 [0.964, 0.997] & 0.995 [0.972, 0.999] & 1.000 [0.981, 1.000] & 79.417 [23.612, 136.648] \\
\midrule
\(\pi/2\) & Reference & 0.900 [0.851, 0.934] & 0.950 [0.910, 0.973] & 0.985 [0.957, 0.995] & 11.764 [11.764, 11.764] \\
 & MIC & 0.920 [0.874, 0.950] & 0.965 [0.930, 0.983] & 0.990 [0.964, 0.997] & 13.029 [12.290, 13.501] \\
 & IMMC & 0.980 [0.950, 0.992] & 0.995 [0.972, 0.999] & 1.000 [0.981, 1.000] & 86.669 [23.751, 137.184] \\
\midrule
\(\pi\) & Reference & 0.890 [0.839, 0.926] & 0.965 [0.930, 0.983] & 0.990 [0.964, 0.997] & 11.764 [11.764, 11.764] \\
 & MIC & 0.900 [0.851, 0.934] & 0.970 [0.936, 0.986] & 1.000 [0.981, 1.000] & 12.929 [12.168, 13.380] \\
 & IMMC & 0.985 [0.957, 0.995] & 0.995 [0.972, 0.999] & 1.000 [0.981, 1.000] & 76.473 [22.353, 138.914] \\
\bottomrule
\end{tabular}
\end{table}

\begin{figure}
    \centering
\includegraphics[width=1\linewidth]{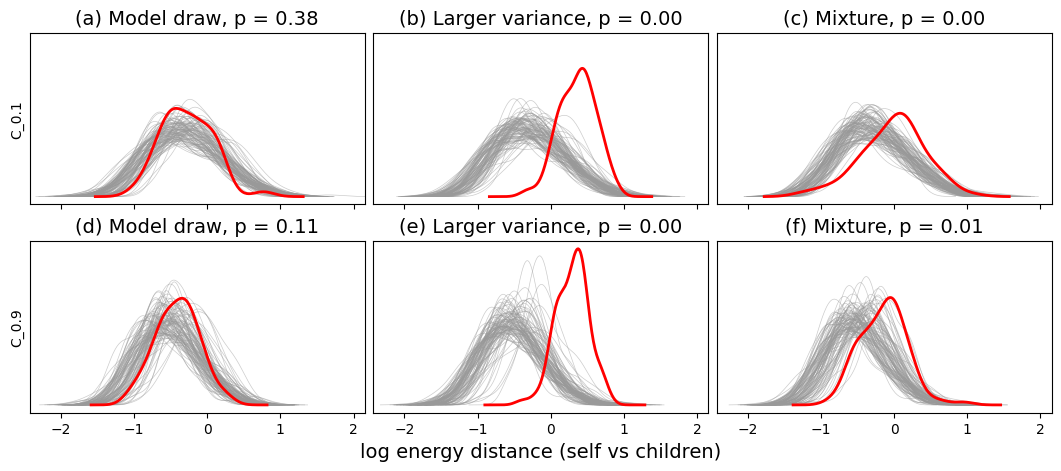}
\caption{Predictive screening for model-generated data ($\sigma = 2.5$; left), inflated deviations ($\times 1.5$;
middle), and a Gaussian mixture with approximately matched marginal variances (right). All
share the same sample mean and contour. Red: $z_0$ to children (predictive) distances. Gray: children to grand children distances. p is a diagnostic score.}
    \label{fig:predictive}
\end{figure}

\textbf{3. Predictive Check.} Figure~\ref{fig:predictive} illustrates our predictive check with the amortized $Q^{\rm MIC}$. We sample children and the grandchildren by using source draws satisfying $\|U\|\leq 1-\alpha$, for $\alpha\in\{0.1,0.9\}$. The gray curves are the KDEs of $\{\log d_{ij}\}_{j=1}^{100}$ for each $i$, providing an internal predictive reference, while the red curve is the KDE of $\{\log d_{0i}\}_{i=1}^{100}$, i.e.\ the log distances from the observed data $z_0$ to its predictive samples. The reported $p$ is the fraction of gray medians that exceed the red median (diagnostic score). As Figure~\ref{fig:predictive} shows, the scale inflation case provides a particularly flattering illustration for our predictive check as it is based on a distance (second column). Notice, however, that the mixture case 
also yields small diagnostic scores despite approximately matched marginal variances,
illustrating a discrepancy beyond inflated marginal spread (third column).

\section{Hypothesis selection with the Bel--Pl Spectrum}
\label{why_Q}
Here, we explain how $Q_z^{\mathrm{MIC}}$ helps us peek into the hypothesis geometry hidden behind the Bel--Pl endpoints. Our goal is to use this geometric information to select among interpretable hypotheses that satisfy a prescribed Bel--Pl decision criterion. Note that the test rule with Pl/Bel in \eqref{eq:decision} gives the equivalence
$P_\theta\{\exists H\ni\theta:\mathrm{Pl}_Z(H)\leq\alpha\}
= P_\theta\{\pi_Z(\theta)\leq\alpha\}$. This equips the test with a uniform validity guarantee \citep{martin2025no, cella2023possibility},
\begin{equation}\label{eq:uniform-validity}
\sup_{\theta\in\mathbb T} P^{(n)}_\theta\bigl\{Pl_{Z_n}(H) \leq\alpha \ \text{ for some measurable } H\ni\theta\bigr\}\leq\alpha.
\end{equation}

Given such a hypothesis probing or tuning is possible, if we are researchers across disciplines analyzing their data, we may be mindful of three things; 1) We may want the accepted hypothesis to be interpretable; an analyst may have a set of interpretable hypotheses, say ${\xi>0}$ or ${\theta_1>\theta_2}$, and may be interested in selecting among these interpretable hypotheses; 2) We may want the accepted hypothesis to be as informative as possible, for example, to have a small area. While $H=C_\alpha$ would be the most efficient, but it may not be interpretable; and 3) When having a similarly sized hypothesis (or union of hypothesis), we may want one that has the largest mean conviction (contour) within the hypothesis.  

\begin{figure}[t]
    \centering
    \includegraphics[width=0.8\linewidth]{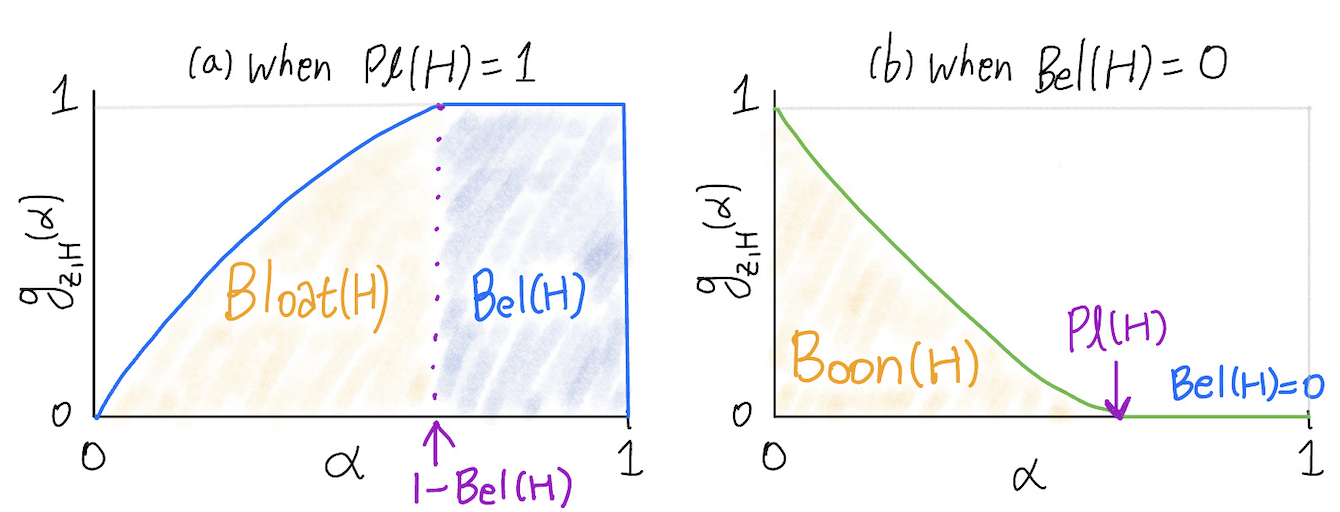}
    \caption{The Bel-Pl spectrum for a hypothesis $H$.}
    \label{fig:bel-pl-spectrum}
\end{figure}

\paragraph{Bel-Pl spectrum.} Our sampler \(Q_z^{\mathrm{MIC}}\) can help to compute useful values for this process providing useful diagnostics values. The first is a diagnostics tool a Bel-Pl spectrum $(Bel(H), \hat{g}_{z,H}(\cdot),$ $ Pl(H))$. This is a useful diagnostics tool for understanding the area efficiency of a hypothesis in terms of contour level set occupancy. By this we can see the gap between the goal of accept/reject and the current hypothesis. We first define $g_{z,H}: \alpha\mapsto[0,1]$, a shellwise allocation function of $H$ as follows. Assuming that the contour shell coincides with $\partial C_\alpha(z)$, we can intuitively write that for any measurable set $H$,
\begin{equation}\label{eq:profile}
g_{z,H}(\alpha):=Q_z^{\rm MIC}\{\Theta\in H \mid \pi_z(\Theta)=\alpha\}
=\frac{|\partial C_\alpha(z)\cap H|}{|\partial C_\alpha(z)|},
\end{equation}
where $|\cdot|$ denotes the declared reference measure on the contour
shell, and the conditional notation refers to the shell kernel ${\rm Unif}(L_{z,\alpha})$ introduced in Section A\footnote{The profile between the endpoints, including its degree of low-depth occupancy, remains reference-measure dependent.}.  A conceptual example of the Bel-Pl spectrum is illustrated in Figure \ref{fig:bel-pl-spectrum}. The Bel--Pl spectrum has a flat right tail at zero or one unless $\mathrm{Bel}_z(H)=0$ and $\mathrm{Pl}_z(H)=1$ simultaneously.\footnote{For shell-regular, full-dimensional $H$ and $H^c$, the spectrum first departs from zero at $\alpha=\mathrm{Pl}_z(H)$ or from one at $\alpha=1-\mathrm{Bel}_z(H)$ as $\alpha$ decreases from one, identifying the nontrivial endpoint.}

\paragraph{Bloat and Boon.} 
The profile gives the proportion of each shell occupied by \(H\). In Figure \ref{fig:bel-pl-spectrum}(a-b), the orange shaded  area measures its accumulated low-depth occupancy. Moving from $\alpha=1$ toward $\alpha=0$, the profile in Figure \ref{fig:bel-pl-spectrum}(a) first departs from one at $\alpha^*=1-\operatorname{Bel}_z(H)$, whereas the profile in Figure \ref{fig:bel-pl-spectrum}(b) first departs from zero at $\alpha^*=\operatorname{Pl}_z(H)$. In Figure \ref{fig:bel-pl-spectrum}(a), when the orange shaded area is large, it motivates searching for a tighter hypothesis within the interpretable class. A small shaded area, on the other hand, indicates little excess occupancy beyond \(C_t(z)\) for $t=1-Bel(H)$. When we are in a rejecting setting like in Figure \ref{fig:bel-pl-spectrum}(b), a larger orange shaded area below the error threshold means more options are rejected, making the test more informative. Since the profile measures proportions rather than absolute size, the area does not mean the actual size of \(C_t(z)\) but the occupancy rate of $H$ on each $\partial C_\alpha$.

\paragraph{Within-hypothesis mean contour depth} While $Q_z^{\rm MIC}(H)$ measures the area under the spectrum, it leaves a vagueness as an efficiency metric: for example, consider two sets, $A=\{\pi_z(\theta)>0.5\}$ and $B=\{\pi_z(\theta)<0.5\}$. They have the same probability $1/2$ w.r.t.\ $Q_z^{\rm MIC}$, while  the contour gives more conviction to $A$. Handling this vagueness, we define the expected contour $\pi$ value,
\begin{equation}\label{eq:expected-contour}
\bar\pi_z(H):=E_{\Theta\sim Q_z^{\rm MIC}}\bigl[\pi_z(\Theta)\mid\Theta\in H\bigr]
=\frac{\int_H \pi_z(\theta)\,Q_z^{\rm MIC}(d\theta)}{Q_z^{\rm MIC}(H)}
=\frac{\int_0^1 \alpha\, g_{z,H}(\alpha)\,d\alpha}{\int_0^1 g_{z,H}(\alpha)\,d\alpha},
\end{equation}
with convention of $\bar\pi_z(H)=0$ when $Q_z^{\rm MIC}(H)=0$. This weighted shell-occupancy profile mean reveals whether that occupancy is concentrated at high or low contour depths. For example, for above sets $A$ and $B$, now we have $\bar\pi_z(A)=3/4$ and $\bar\pi_z(B)=1/4$. Besides, dividing by $Q_z^{\rm MIC}(H)$ prevents a set from being favored simply because it has high mass.
It is easy to show given $H_1$ and $H_2$ accepted at level \(\alpha\in(0,1/2)\), if $H_1\subseteq H_2$ and
$Q_z^{\rm MIC}(H_2\setminus H_1)>0$, then $\bar\pi_z(H_1)>\bar\pi_z(H_2)$. This means when we have multiple candidate hypothesis that has inclusion relationship, a smaller set will be prioritized based on $\bar\pi_z$. 

\paragraph{Hypothesis Selection or Rejection.} Given surplus measures such as Bloat and Boon, along with the expected depth, there are many ways to formulate a hypothesis selection or composition rule. The optimal criterion would depend on the analyst's specific context. Here, we suggest one of many possible selection rules. Putting these ingredients together, we define the hypothesis selection rule
\begin{equation}\label{eq:Bel_opt}
\begin{aligned}
    \widehat H(z)\in&\operatorname*{arg\,max}_{H\in\mathcal H}\quad
    \mathbb E_{Q_z^{\mathrm{MIC}}}
    \bigl[\pi_z(\Theta)\mid\Theta\in H\bigr]
    \\
    &\text{subject to}\quad
    \mathrm{Bel}_z(H)\ge 1-\alpha_0 \text{ and }
   Q_z^{\rm MIC}(H)-Bel_z(H)\le\beta,
\end{aligned}
\end{equation}
where $\mathcal H$ is a class of interpretable candidate hypotheses,
$\alpha_0\in(0,1)$ is a fixed inferential error level, and $\beta\ge0$
bounds the intrinsic bloat $Q_z^{\mathrm{MIC}}(H)-\mathrm{Bel}_z(H)$.
Since $\mathrm{Bel}_z(H)\ge1-\alpha_0$ implies
$g_{z,H}(\alpha)=1$ for almost every $\alpha\in(\alpha_0,1)$,
the optimization compares the contour depth and amount of the
excess occupancy outside the shared core $C_z(\alpha_0)$.

In the rejection setting, we could seek to exclude more MIC mass, giving greater weight to lower contour values.
For example,
we could select
\begin{equation}\label{eq:when_rejecting}
\begin{aligned}
\widehat H(z)\in
\operatorname*{arg\,max}_{H\in\mathcal H}\quad&\int_H \{1-\pi_z(\theta)\}\,Q_z^{\mathrm{MIC}}(d\theta)
\quad \text{subject to}\quad \mathrm{Pl}_z(H)\le\alpha_0,
\end{aligned}
\end{equation}
which is equivalent to maximizing $Q_z^{\mathrm{MIC}}(H)\,
\mathbb E_{Q_z^{\mathrm{MIC}}}
\bigl[1-\pi_z(\Theta)\mid\Theta\in H\bigr]$.

\begin{figure}
    \centering
    \includegraphics[width=1\linewidth]{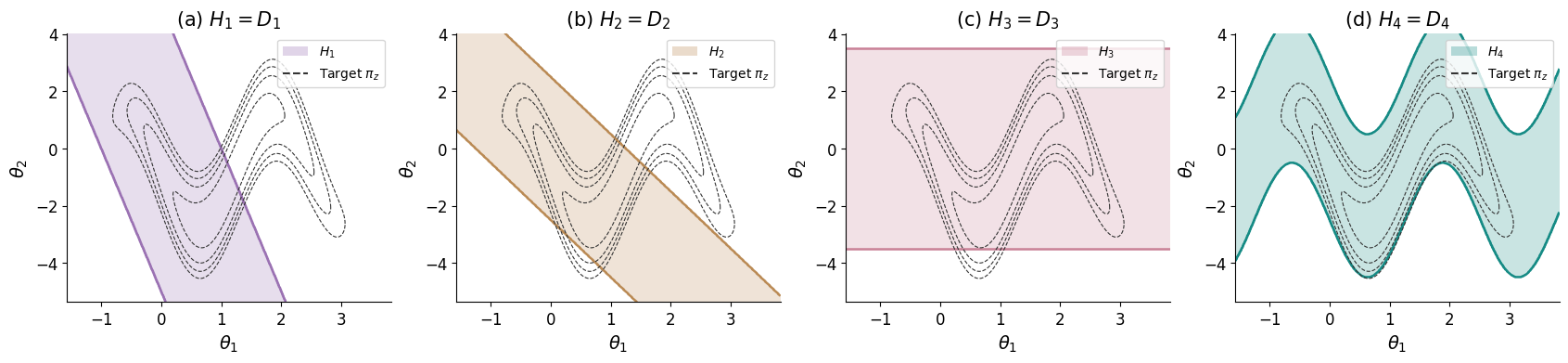}
    \caption{The primal candidate hypotheses in Example 3. Their 15 nonempty unions are considered as candidates.}
    \label{fig:candidates}
\end{figure}
\begin{figure}
    \centering
    \includegraphics[width=.9\linewidth]{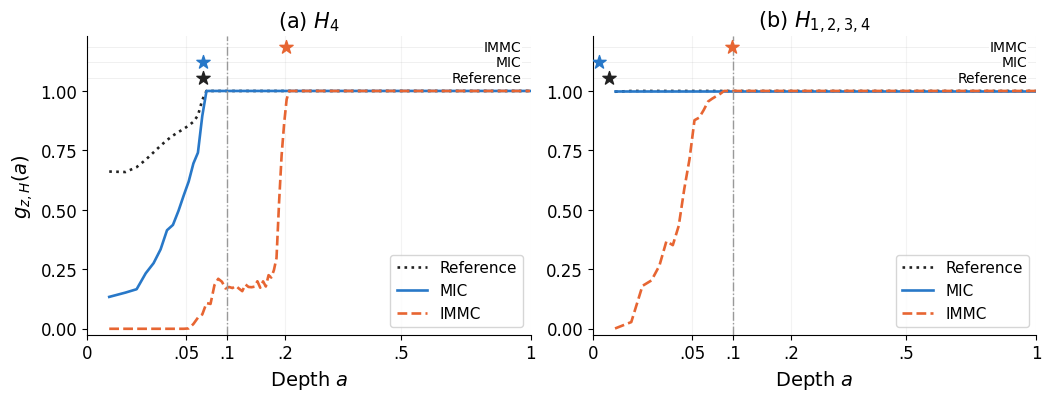}
    \caption{The Bel-Pl spectrum comparison for the selected hypothesis. (a) MIC and the reference selected $H_4$, while IMMC estimates showed $H_4$ as not meeting the constraint. (b) IMMC selected $H_{1,2,3,4}=\cup_{i=1}^4 H_i$. Stars located at the estimated $1-Bel(H_\cdot)$.}
    \label{fig:selected}
\end{figure}

\paragraph{Example 3.}
We continue in the Shear data. The basis candidate hypotheses are $D_1=\{|\theta_2+5\theta_1|\le 5\}$,
$D_2=\{|\theta_2+2\theta_1|\le 2.5\}$, $D_3=\{|\theta_2|\le 3.5\}$, and
$D_4=\{|\theta_2+2\sin(2.5\theta_1)|\le 2.5\}$ as in Figure \ref{fig:candidates}. The 15 nonempty unions $H_S=\bigcup_{i\in S}D_i$, $\varnothing\neq S\subseteq\{1,2,3,4\}$, are considered as candidates in the selection according to \eqref{eq:Bel_opt} with $\alpha_0=0.10$ and $\beta=0.30$. The selection results are in Figure \ref{fig:selected}. As shown, the IMMC estimate of $Bel(H_4)$ (the orange star) is much smaller than that of MIC and the reference (the blue and black stars). While MIC and the reference selected $H_4$, IMMC selected $H_{1,2,3,4}=\cup_{i=1}^4 H_i$ as the most efficient based on \eqref{eq:Bel_opt}.

\section{Application to Ovarian Aging}\label{sec:ovarian}

The real-life problem we consider here is ovarian aging. The number of women working with their own expertise has continued growing, far beyond what it was a century ago. As it has become common for the biological window for having children to overlap with the time needed to develop and establish a career, knowing how much ovarian reserve one has left can play a very important role in life planning \citepalias{asrm2024planned}. Freezing eggs at a younger age and using them for IVF when desired is also widely considered, but the cost, time, concerns about side effects, and personal values mean that this is not an easily accessible path for everyone \citepalias{asrm2024planned}. 

Women are born with 1--2 million eggs, and this number declines over time, leaving only around 1,000 (about 0.1\%) by menopause \citepalias{acog2014decline}. But both the rate of loss and the number present at birth differ from person to person although there are patterns across age groups \citep{wallace2010human,de2016back}. AMH is a hormone released by small follicles in the ovaries that have just begun to grow, and is currently used as an indirect proxy for the remaining ovarian reserve \citetext{\citealp{steiner2017association}; \citetalias{asrm2020reserve}}.

We analyze synthetic records constructed to match the age-specific log-AMH curve and between-individual variation reported for the Doetinchem cohort in \cite{de2016back}. Our model is as follows. Let $Y^*_{ij}$ be the latent log-AMH value measured for woman $i$ at age $t_{ij}$. Then
\begin{equation}\label{eq:amh_model}
Y^*_{ij} = f(t_{ij} + c_i) + a_i + \varepsilon_{ij}, \qquad
\varepsilon_{ij} \overset{\text{iid}}{\sim} N(0, \sigma^2),
\end{equation}
for $c_i\in[-15, 15], ~a_i\in [-4, 2],~\sigma = 0.3,$ and $j = 1,\dots,4.$
Here, $f$ is the population median log-AMH curve over age, and we take the curve estimated from a large cohort as given. There are two individual parameters in this model. The parameter $c_i$ answers the question, ``How many years ahead or behind is my AMH trajectory relative to the population reference?'' The parameter $a_i$ captures individual differences in log-AMH levels after accounting for this age shift. AMH measurements have a detection limit, so we treat measurements below this limit as left-censored observations. See the appendix for details of the data, model, and parameter ranges, along with their rationale. 

Our goal is to use IM to make inferences for each person under this model. In particular, we want to say how many more years this person's median AMH will remain above a reference value $b$. Let $t_*$ be the age at the last measurement. The hypothesis that ``it will still be above $b$ after $k$ years'' is
\begin{equation}
H_k :=\{(c,a) : f(t_*+k+c) + a \ge \log b\}, \qquad k = 0, 1, 2, \dots
\end{equation}
Since $f$ is decreasing, these hypotheses are nested as $H_0 \supseteq H_1 \supseteq H_2 \supseteq \cdots$. The question then becomes, ``Up to which $k$ can we say this with confidence?'' We report the largest $k$ satisfying $\mathrm{Bel}_z(H_k) \ge 1-\alpha$ as ``at least $k$ years remain'' for this person. Conversely, if there is a smallest $k$ satisfying $\mathrm{Pl}_z(H_k) \le \alpha$, we report that ``fewer than $k$ years remain.'' For values of $k$ that meet neither condition, we withhold judgment. We set the reference value to $b = 0.946$ ng/mL as justified in the appendix. We focus on the 160 women (hypothetical) whose MLEs imply a positive remaining time.

\begin{figure}
    \centering
    \includegraphics[width=1\linewidth]{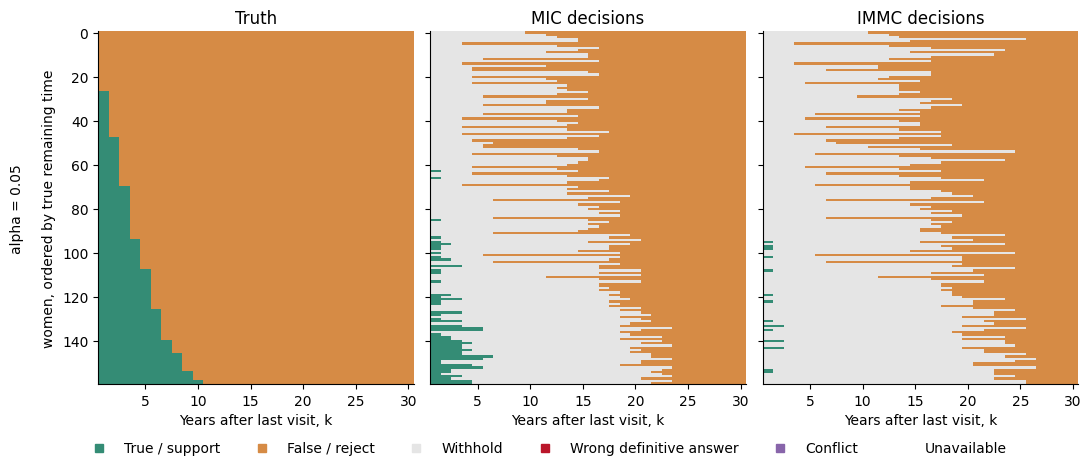}
    \caption{Left: The true remaining years for which median AMH stays above the threshold. Center: Inference results from MIC. Right: Inference results from IMMC. A larger green area (closer to the truth) means more informative inference.}
    \label{fig:decisions}
\end{figure}
\begin{figure}
    \centering
    \includegraphics[width=1\linewidth]{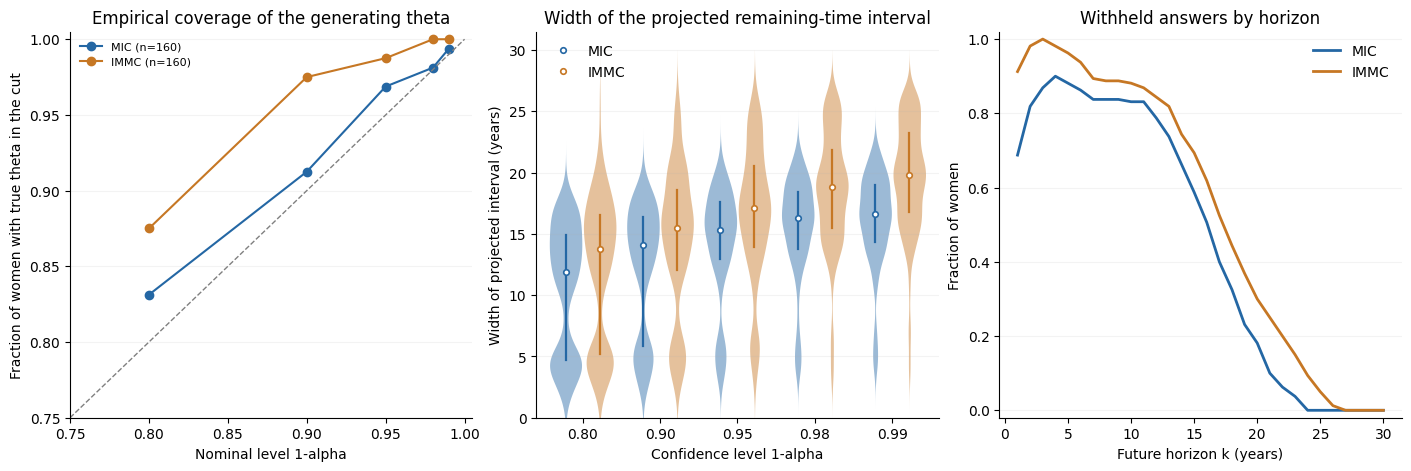}
    \caption{MIC yields tighter $\alpha$-cuts. Left: Empirical coverage of $C_\alpha$ for the true parameter $\theta$ across 160 synthetic women. Middle: The widths of the projected remaining-time intervals. Right: Fraction of women for whom judgment is withheld at each future horizon, with $\alpha=0.05$.}
    \label{fig:coverage}
\end{figure}

We wish that we could work with the real data in \citep{de2016back} rather than the synthetic data constructed by digitizing their reported results. (The appendix also shows how our synthesized data shows similar summary behavior as the data reported in \cite{de2016back}.)
Thankfully, however, we know the true generating parameters as we use the synthetic data. This means that we also know each person's true remaining duration (synthesized), thus enabling us to evaluate the empirical coverage.\footnote{Once our model's performance has been well validated on synthetic data, we can apply the same analysis again when actual longitudinal records become available.} We first fix $\alpha=0.05$ and try the hypothesis probing that interests us most. In this context, this simply means finding the largest $k$ satisfying $Bel(H_k)\geq 0.95$. For larger values of $k$, we can also reject those for which $Pl(H_k)\leq 0.05$. The results are in Figure \ref{fig:decisions} and Figure \ref{fig:MC_ovarian} in Appendix. As we can see, MIC inferred more additional time for more women (larger green area), and also ruled out a larger range of remaining durations (larger orange area), and higher consistency with the MC target. As shown in Figure \ref{fig:coverage}, we can see that MIC comes closer to the nominal coverage level and gives narrower confidence intervals. As a result, the withhold rate also decreases. See Table \ref{tab:amh_decision_details} in Appendix for more detailed decision comparisons. For example, at $\alpha=0.05$, the withhold rate, averaged over the 160 women and all evaluated horizons, is 46.0\% for MIC and 54.4\% for IMMC. Across these horizons, no incorrect definitive judgments occur MIC, IMMC, or the MC reference.

Finally, we look at the data for a few (hypothetical) women. Figure \ref{fig:amh_spectrum} in Appendix shows how various hypotheses were accepted or rejected for each woman (each row). As we can see in the left column, MIC's $\alpha$-cuts tend to be slightly tighter. Consequently, the right column shows cases where MIC can reject or accept hypotheses that IMMC cannot.

\section{Concluding Remarks}\label{sec:discuss}
We developed a radial transport formulation for learning an MIC sampler without first approximating contour-cut boundaries. Amortization allows this sampler to be reused across datasets, while the Bel--Pl spectrum uses its shellwise information to compare and select interpretable hypotheses. In our experiments, the learned contours and resulting coverage, power, and hypothesis selections are close to the reference results. In the synthetic AMH analysis, MIC gives decisions closer to the MC reference and fewer withheld judgments than the baseline.

There are several reasons why we call our method a ``junior AI''. Large AI systems such as ChatGPT, Claude, and Gemini take substantial time and cost to train, but once trained, they produce answers relatively quickly. Likewise in our case, training the network can take time, but once well trained and validated, the network can be used repeatedly on new datasets without additional training. We call it ``junior'' because its competence is limited to a particular sample size $n$ and a particular likelihood model.

This initial investment of training can pay off when many datasets or many hypotheses must be analyzed, as in automated scientific discovery. For example, suppose thousands of protein variants or vaccine candidates can be evaluated under a common model, and we want to know which of them satisfy a specified statistical criterion. If an amortized inferential machine is already available on the device or a remote server, each new measurement need not trigger another lengthy computation. Likewise, when there is little time to wait, such a junior AI may shine. For example, how long would people be willing to wait for a statistical assessment of the uncertainty behind an alert from a wearable device?

In our work, we established  exact results for the radial representation of inner credal distributions and the maximum-entropy characterization (Theorems~1 and~2). On the other hand, our learned transport is an approximation, for which we do not establish finite-sample error bounds in the current version of the manuscript. Instead, amortization allows us to assess the learned sampler’s coverage and power through extensive simulation before it meets real data. We can flag a new dataset whose relationship with its predictive children is quite different from what may be thought as typical from the sampler's perspective.

Using deep learning means there are hyperparameters to tune, such as the learning rate, optimizer, training schedule, and network architecture. While one may begin from our default settings, harder problems may call for different ones. In our algorithm, an important hyperparameter is $\gamma_h$, the weight of the entropy loss. While we fixed $\gamma_h = 0.01$ across the examples for simple presentation, systematic selection of this weight may offer further gains in calibration.

\bibliographystyle{chicago}
\typeout{}
\bibliography{ref}

\appendix

\startcontents[appendix]
\printcontents[appendix]{l}{1}{\section*{Appendix Table of Contents}}

\section{Proof of Theorems}\label{sec:proof}

In the IM framework, the possibilistic quality of a parameter $\theta$ is measured through the IM contour $\pi_{z_n}(\theta)$. While membership in $\mathcal{C}^{\rm inn}(\Pi)$ fixes the marginal distribution of contour depth, it leaves the conditional distributions within level sets unspecified.

For $\alpha\in(0,1)$, define a contour-based level set
$L_{z_n,\alpha}:=\{\theta\in\mathbb{T}:\pi_{z_n}(\theta)=\alpha\}.$
For $L_{z_n,\alpha}$ s.t.
$0<\operatorname{Area}(L_{z_n,\alpha})<\infty,$%
\footnote{Here, $\operatorname{Area}$ denotes Euclidean
$(d-1)$-dimensional surface area.}
we define uniform distribution on $L_{z_n,\alpha}$:
For every Borel set $B\subseteq\mathbb T$, define
\[
\operatorname{Unif}(L_{z_n,\alpha})(B)
=
\frac{
\operatorname{Area}(B\cap L_{z_n,\alpha})
}{
\operatorname{Area}(L_{z_n,\alpha})
},
\]
where uniformity and entropy are understood relative to a declared shell reference measure, taken here to be the Euclidean $(d-1)$-dimensional surface measure in the chosen parameterization. Here now on, let $H_{z_n,\alpha}$ denote entropy with respect to Euclidean
surface area on $L_{z_n,\alpha}$%
\footnote{Specifically, for any probability
distribution $K$ on $L_{z_n,\alpha}$ that is absolutely continuous
with respect to surface area,
\[
H_{z_n,\alpha}(K)
=
-\int_{L_{z_n,\alpha}}
\log\!\left(
\frac{dK}{d\operatorname{Area}}(\theta)
\right)K(d\theta).
\]
If $K$ is not absolutely continuous with respect to surface area, we
set $H_{z_n,\alpha}(K)=-\infty$.
}.

\subsection{Proof of Proposition~\ref{prop:elliptical-distortion}}
\label{sec:proof-elliptical-distortion}

Fix $\varepsilon\in(0,1/2)$.
We first construct a specific counterexample, a model whose cuts become thin while their minimum-area MLE-centered enclosing ellipses remain fixed.
For $\kappa>0$ (to be chosen below), consider a single observation
\[
Z\sim N_2(m_\kappa(\theta),I_2),
\qquad
\theta=(x,y),
\qquad
m_\kappa(x,y)=\bigl(x,\,2(1+\kappa x^2)y\bigr).
\]
Note that $m_\kappa$ is a smooth map of $\mathbb R^2$ onto itself.
Thus $\widehat\theta_z=m_\kappa^{-1}(z)$, and the normalized likelihood is
\[
R_z(\theta)
=\exp\!\left\{-\frac12\lVert z-m_\kappa(\theta)\rVert^2\right\}.
\]
Under $P_\theta$, we have $\lVert Z-m_\kappa(\theta)\rVert^2\sim\chi_2^2$, so $R_Z(\theta)\sim\Unif(0,1)$.
By \eqref{eq:im_contour}, the exact
IM contour thus satisfies $\pi_z(\theta)=R_z(\theta)$.
Take $z=0$, for which $\widehat\theta_0=0$. Then we can write the true contour as
\[
\pi_0(x,y)
=\exp\!\left[-\frac{x^2+4(1+\kappa x^2)^2y^2}{2}\right].
\]
For $a\in(0,1)$, let $r_a=\sqrt{-2\log a}$. The cut
$C_a(0)=m_\kappa^{-1}\{\eta:\lVert\eta\rVert\le r_a\}$ is compact and
connected, with a smooth boundary, due to the smoothness of the map $m_\kappa$.
For each fixed $a$, increasing $\kappa$ makes $C_a(0)$ increasingly thin around the coordinate axes, while its four axis intercepts remain fixed.
The cut therefore approaches a cross formed by two line segments along the coordinate axes.

We next show that the minimum-area MLE-centered enclosing ellipses remain fixed as $\kappa$ increases.
Since $1+\kappa x^2\ge1$,
\[
C_a(0)\subseteq C_a^{\mathrm{out}}(0)
:=\{(x,y):x^2+4y^2\le r_a^2\}.
\]
The cut contains the four points $(\pm r_a,0)$ and $(0,\pm r_a/2)$.
Although the cut becomes increasingly thin, every enclosing ellipse must still contain these four fixed points.
Yet these points determine the minimum-area MLE-centered enclosing ellipse.
Any centered enclosing ellipse $\{\theta:\theta^\top B\theta\le1\}$, with $B$ symmetric positive definite, must therefore satisfy
\[
B_{11}\le r_a^{-2},
\qquad
B_{22}\le4r_a^{-2},
\qquad
\det B\le B_{11}B_{22}\le4r_a^{-4}.
\]
Its area is at least $\pi r_a^2/2$, which is attained by $C_a^{\mathrm{out}}(0)$.
Equality forces
$B=\operatorname{diag}(1,4)/r_a^2$.
Thus $C_a^{\mathrm{out}}(0)$ is the unique minimum-area MLE-centered enclosing ellipse.

For independent $A\sim\Unif(0,1)$ and $V\sim\Unif(\mathbb S^1)$, the population IMMC law is the distribution of $(r_A V_1, r_A V_2/2) \stackrel{d}{=} (X,Y/2)$, with $X,Y \overset{\mathrm{iid}}{\sim} N(0,1)$,
and the equality in distribution follows from $r_A^2\sim\chi_2^2$.
Therefore $Q_0^E=N_2(0,\operatorname{diag}(1,1/4))$ for every $\kappa>0$.

We now show that the population stitched contour based on $Q_0^{\mathrm{MIC}}$ recovers the exact IM contour.
We use arclength as the shell reference measure. The smooth family of compact level curves defines a measurable probability kernel $a\mapsto\nu_{0,a}$.
Note that each quadrant of $\partial C_a(0)$ is a monotone graph joining its two axis intercepts.
Therefore
\[
4r_a\le |\partial C_a(0)|\le6r_a,
\qquad
\int_0^1\bigl|\log|\partial C_a(0)|\bigr|\,\mathrm da<\infty,
\]
where the integrability follows from $r_a=\sqrt{-2\log a}$.
Thus the shell reference measure satisfies the entropy integrability condition in Theorem~\ref{thm:entropy_character}.
By the definition of $Q_0^{\mathrm{MIC}}$ in \eqref{eq:def_MIC},
$\pi_0(\Theta)\sim\Unif(0,1)$ under this law.
Thus, for every contour depth $t\in(0,1)$, this distribution assigns probability $t$ to parameter values whose exact IM contour values do not exceed $t$.
Meanwhile, since $R_0=\pi_0$, the population stitched contour satisfies
\[
\rho_0^{\mathrm{MIC}}(\theta)
=Q_0^{\mathrm{MIC}}
\{\pi_0(\Theta)\le\pi_0(\theta)\}
=\pi_0(\theta).
\]

We finally construct a hypothesis whose plausibility is arbitrarily small under the exact IM contour but arbitrarily close to one under the population stitched contour based on $Q_0^E$.
Consider the half-space
\[
H=\{(x,y):y\ge r_\varepsilon/2\}.
\]
Note that $\pi_0(x,y)\le e^{-2y^2}\le\varepsilon$ on $H$, with equality at $(0,r_\varepsilon/2)$. Since $R_0=\pi_0$, the population stitched contour is nondecreasing in $\pi_0(\theta)$.
Thus
\[
\operatorname{Pl}_{\pi_0}(H)=\varepsilon,
\qquad
\operatorname{Pl}_{\rho_0^E}(H)
=Q_0^E\{\pi_0(\Theta)\le\varepsilon\}.
\]
As $\kappa$ increases, $Q_0^E$ remains fixed while the true contour cuts contract.
Consequently, an increasing proportion of its draws have exact IM contour values below $\varepsilon$, making the stitched plausibility of $H$ close to one.
In fact, using the independence of $X$ and $Y$, the bound
$\Pr(|Y|<b)\le\sqrt{2/\pi}\,b$ for $b\ge0$, and the bound
$(2\pi)^{-1/2}$ on the standard normal density of $X$, we obtain
\begin{align*}
1-\operatorname{Pl}_{\rho_0^E}(H)
&=\Pr\{X^2+(1+\kappa X^2)^2Y^2<r_\varepsilon^2\}\\
&\le
\Pr\!\left(
|Y|<\frac{r_\varepsilon}{1+\kappa X^2}
\right)\\
&\le
\sqrt{\frac{2}{\pi}}\,
r_\varepsilon\,
\E\!\left[\frac{1}{1+\kappa X^2}\right]\\
&\le
\frac{r_\varepsilon}{\pi}
\int_{\mathbb R}\frac{\mathrm dx}{1+\kappa x^2}
=\frac{r_\varepsilon}{\sqrt{\kappa}}.
\end{align*}
Choosing $\kappa\ge r_\varepsilon^2/\varepsilon^2 = (-2\log\varepsilon)/\varepsilon^2$ therefore gives $\operatorname{Pl}_{\rho_0^E}(H) \ge 1-\varepsilon$.
This proves the proposition.

\subsection{Proof of Theorem~\ref{thm:main-radial}}\label{app:pf.main-radial}

We first show that every inner radial transport generates an inner credal distribution.
With fixed $z$, for $T\in\mT_{\mathrm{in}}(\pi_z)$, the depth constraint gives
\[
\pi_z\{T(U)\}=1-\lVert U\rVert\sim\Unif(0,1).
\]
Thus $(T_\#F_U)(C_\alpha(z))=1-\alpha$ for every $\alpha\in[0,1]$.
Moreover, for any nonempty Borel set $B\subseteq\TT$, the event $T(U)\in B$ implies
$\pi_z\{T(U)\}\le\Pi_z(B)$.
Therefore
\[
(T_\#F_U)(B)
\le \Pr\bigl\{\pi_z\{T(U)\}\le\Pi_z(B)\bigr\}
=\Pi_z(B).
\]
The inequality also holds for $B=\varnothing$. Hence
$T_\#F_U\in\mathcal C^{\mathrm{inn}}(\Pi_z)$.

Conversely, we show that every inner credal distribution can be generated by an inner radial transport.
The source radius determines the contour depth, while the independent direction provides the randomness needed to sample within the corresponding shell.
Take $Q\in\mathcal C^{\mathrm{inn}}(\Pi_z)$.
By \eqref{eq:calibration}, $\pi_z(\Theta)\sim\Unif(0,1)$ under $Q$.
Since $\TT$ is standard Borel and $\pi_z$ is measurable, there is a regular conditional probability kernel $K^a = Q(\,\cdot \mid \pi_z(\Theta)=a)$ such that
\[
Q(B)=\int_0^1 K^a(B)\,\d a,
\qquad
K^a(L_{z,a})=1
\quad\text{for almost every }a,
\]
for every Borel set $B\subseteq\TT$.

We next construct a sampler for the conditional distributions $K^a$.
By the randomization lemma for probability kernels on standard Borel spaces, there exists a jointly measurable map $G:[0,1]^2\to\TT$ such that $G(a,W)\sim K^a$ whenever $W\sim\Unif(0,1)$.
Note that $d\ge2$ ensures that the first coordinate $V_1$ of $V\sim\Unif(\mathbb S^{d-1})$ has a continuous CDF $F_{V_1}$.
Thus the source direction supplies the required uniform variable:
\[
W=F_{V_1}(V_1)\sim\Unif(0,1),
\qquad W\perp A.
\]
Fix any $\theta_*\in\TT$ and define
\[
T(u)=
\begin{cases}
G\!\left(1-\lVert u\rVert,
          F_{V_1}\!\left(u_1/\lVert u\rVert\right)\right),
   &0<\lVert u\rVert<1,\\
\theta_*,&\text{otherwise}.
\end{cases}
\]
This map is measurable, and the second case concerns only an $F_U$-null set. Since $U=(1-A)V$, we have $T(U)=G(A,W)$ almost surely.
By the independence of $A$ and $W$, the conditional law of $T(U)$ given $A=a$ is $K^a$ for almost every $a$.
Therefore
\[
(T_\#F_U)(B)=\int_0^1 K^a(B)\,\d a=Q(B)
\]
for every Borel set $B\subseteq\TT$.
Moreover, since $K^a(L_{z,a})=1$ for almost every $a$,
\[
\pi_z\{T(U)\}=A=1-\lVert U\rVert
\quad F_U\text{-almost surely}.
\]
Thus $T\in\mT_{\mathrm{in}}(\pi_z)$ and $T_\#F_U=Q$.
Together with the first part of the proof, this establishes assertion (i).

We next show that conditioning the source radius produces the conditional distribution on a cut.
For every $T\in\mT_{\mathrm{in}}(\pi_z)$ and $\alpha\in(0,1)$, the depth constraint implies
\[
\{\lVert U\rVert\le1-\alpha\}
=\{\pi_z(T(U))\ge\alpha\}
=\{T(U)\in C_\alpha(z)\}
\quad\text{up to an }F_U\text{-null set}.
\]
Thus restricting the source radius selects exactly the draws whose images lie in $C_\alpha(z)$.
These events have probability $1-\alpha>0$.
For every Borel set $B\subseteq\TT$, we obtain
\begin{align*}
\left(T_\#\left[
F_U(\,\cdot\mid\lVert U\rVert\le1-\alpha)
\right]\right)(B)
&=
\frac{\Pr\{T(U)\in B,\ \lVert U\rVert\le1-\alpha\}}
     {1-\alpha}\\
&=
\frac{(T_\#F_U)(B\cap C_\alpha(z))}{1-\alpha}\\
&=(T_\#F_U)(B\mid C_\alpha(z)).
\end{align*}
This proves assertion (ii).

Finally, we apply the representation to the maximum-entropy inner credal distribution.
The measurable probability kernel $a\mapsto\nu_{z,a}$ makes $Q_z^{\mathrm{MIC}}$ in \eqref{eq:def_MIC} a probability measure.
Since $\nu_{z,a}(L_{z,a})=1$ for almost every $a$, for every Borel set $D\subseteq[0,1]$,
\[
Q_z^{\mathrm{MIC}}\{\pi_z(\Theta)\in D\}
=\int_0^1\nu_{z,a}\{\pi_z(\Theta)\in D\}\,\d a
=\int_0^1\mathbf 1_D(a)\,\d a.
\]
Thus $\pi_z(\Theta)\sim\Unif(0,1)$ under $Q_z^{\mathrm{MIC}}$.
The same cut-probability and domination arguments used above give $Q_z^{\mathrm{MIC}}\in\mathcal C^{\mathrm{inn}}(\Pi_z)$.
By (i), there exists $T^{\mathrm{MIC}}\in\mT_{\mathrm{in}}(\pi_z)$ such that $(T^{\mathrm{MIC}})_\#F_U=Q_z^{\mathrm{MIC}}$.
This proves the final assertion.

\subsection{Proof of Theorem~\ref{thm:entropy_character}}\label{app:pf.entropy_character}

We first show that fixing the source radius gives the conditional distribution on the corresponding contour shell.
Fix $T\in\mT_{\mathrm{in}}(\pi_z)$. For $a\in(0,1)$, write
\[
K^a :=T_\#\left[
F_U(\,\cdot\mid\lVert U\rVert=1-a)
\right]
=\operatorname{Law}\{T((1-a)V)\}.
\]
The independence of the source radius and direction gives this version of the regular conditional distribution, and $a\mapsto K^a$ is a measurable probability kernel.
By the depth constraint and Fubini's theorem, $K^a(L_{z,a})=1$ for almost every $a$.
Moreover,
\[
(T_\#F_U)(B)=\int_0^1 K^a(B)\,\d a
\]
for every Borel set $B\subseteq\TT$.
Thus $K^a$ is also a version of the conditional distribution of $T(U)$ given $\pi_z\{T(U)\}=a$.
Likewise, by \eqref{eq:def_MIC}, $\nu_{z,a}$ is the conditional distribution of $Q_z^{\mathrm{MIC}}$ on the corresponding shell.

We next express the entropy gap on each shell as a KL divergence.
Let $\mu_{z,a}$ denote the declared shell reference measure.
For almost every $a$, this measure has finite, positive mass, so
\[
\nu_{z,a}=\frac{\mu_{z,a}}{\mu_{z,a}(L_{z,a})},
\qquad
H_{z,a}(\nu_{z,a})=\log\mu_{z,a}(L_{z,a}).
\]
If $K^a\ll\mu_{z,a}$, then
\begin{align*}
H_{z,a}(\nu_{z,a})-H_{z,a}(K^a)
&=\log\mu_{z,a}(L_{z,a})
  +\int_{L_{z,a}}\log\!\left(\frac{\d K^a}{\d\mu_{z,a}}\right)\,\d K^a\\
&=\int_{L_{z,a}}\log\!\left(\frac{\d K^a}{\d\nu_{z,a}}\right)\,\d K^a\\
&=\KL(K^a\Vert\nu_{z,a}).
\end{align*}
If $K^a$ is not much smaller than $\mu_{z,a}$, both sides equal $+\infty$ under the stated entropy convention.
Thus the identity holds in the extended-real sense, including when the KL divergence is infinite.

We now integrate over contour depths to obtain the entropy-gap identity.
Both $T_\#F_U$ and $Q_z^{\mathrm{MIC}}$ have the same uniform contour-depth marginal.
Since contour depth is a measurable function of the parameter, the chain rule for relative entropy gives
\begin{align*}
\KL(T_\#F_U\Vert Q_z^{\mathrm{MIC}})
&=\int_0^1\KL(K^a\Vert\nu_{z,a})\,\d a\\
&=\int_0^1
\left[H_{z,a}(\nu_{z,a})-H_{z,a}(K^a)\right]\,\d a.
\end{align*}
The common contour-depth marginal contributes zero to the KL divergence, so only the differences within shells remain.
The equality also holds when the KL divergence is infinite.
Substituting the definition of $K^a$ proves assertion (i).

We finally show that the integrated conditional entropy is maximized exactly by transports that generate $Q_z^{\mathrm{MIC}}$.
The integrability assumption makes $\int_0^1 H_{z,a}(\nu_{z,a})\,\d a$ finite.
Since $H_{z,a}(K^a)\le H_{z,a}(\nu_{z,a})$ almost everywhere, $\int_0^1 H_{z,a}(K^a)\,\d a$ is well defined in $[-\infty,\infty)$.
Therefore assertion (i) yields
\begin{align*}
\int_0^1 H_{z,a}(K^a)\,\d a
&=\int_0^1 H_{z,a}(\nu_{z,a})\,\d a
  -\KL(T_\#F_U\Vert Q_z^{\mathrm{MIC}})\\
&\le\int_0^1 H_{z,a}(\nu_{z,a})\,\d a.
\end{align*}
By Theorem~\ref{thm:main-radial}, there exists an inner radial transport that generates $Q_z^{\mathrm{MIC}}$, so this upper bound is attained.
Moreover, equality holds if and only if $\KL(T_\#F_U\Vert Q_z^{\mathrm{MIC}})=0$, or equivalently, $T_\#F_U=Q_z^{\mathrm{MIC}}$.
This proves assertion (ii).
Thus the maximizing distribution is uniquely $Q_z^{\mathrm{MIC}}$, although its corresponding transports need not be unique.

\section{Algorithms}\label{sec:algs}
Here, we present our two main algorithms non-amortized version, amortized version, and the training data generating algorithm for the amortized training. 
\vspace{-1cm}
\begin{algorithm}
\caption{MIC Sampler Learning.}
\label{alg:mic}
\spacingset{1.2}
\textbf{Input:} data $z$, rank function $R$.

\textbf{Tuning parameters:} loss weights $\gamma_{h},\gamma_f$, and temperature $(\tau)$.

\textbf{Learnable objects:} transport map $T_\psi$.

\textbf{Data:} $D_{\rm target}=\{(\vartheta_i, \pi(\vartheta_i))\}_{i=1}^{N_{\rm target}}$, $\vartheta_i\sim \operatorname{Unif}(\mathbb{T})$ computed for rank function $R$.

\noindent\colorbox[gray]{0.9}{\parbox{\dimexpr\linewidth-2\fboxsep\relax}{\centering\textbf{Transport Map Learning}}}

\textbf{for $t =1,...,N_{\mathrm{iter}}$ do}

 \quad \quad Sample $A_i\sim \operatorname{Unif}(0,1)$ and $V_j\sim \operatorname{Unif}(\mathbb{S}^{d-1})$.
 
\quad \quad  Define $U_{ij} = (1-A_i)V_j$ for $i\in[N]$ and $j\in[M]$.

\quad \quad  Compute $\widetilde{\Theta}_{ij}=T_\psi(U_{ij})$

\quad \quad  \textbf{Step 1.} Sample particles $U_1,\ldots,U_K\sim F_U$, and transport by $\widetilde\Theta_k=T_\psi(U_k)$.

\quad \quad Compute empirical loss $\mathcal{L}_{d} (\psi)= \frac{1}{MN}\sum_{i=1}^N \sum_{j=1}^M S (A_i,\widehat\rho_t(\widetilde\Theta_{ij}))$, where  \[\widehat\rho_{t}(\widetilde\Theta_{ij})=
  \frac{1}{K}
  \sum_{k=1}^K
  \sigma\!\left(
    \frac{R(\widetilde\Theta_{ij})-R(\widetilde\Theta_k)}{\tau}
  \right).\]

  \quad \quad   \textbf{Step 2.} Compute a shell 1-nn distance by $R_{ij}^\psi = \min_{t\neq j} \| \widetilde{\Theta}_{ij}-\widetilde{\Theta}_{it}\|$

\quad \quad Compute empirical loss 
\vspace{-7mm}
\[\mathcal{L}_{h}(\psi)=-\frac{d-1}{NM}\sum_{i=1}^N \sum_{j=1}^M \log R_{ij}^\psi,\]

  \quad \quad     \textbf{Step 3.} Sample $(\vartheta_1,\pi(\vartheta_1)),...,(\vartheta_K,\pi(\vartheta_K))$ from $D_{\rm target}$. Compute 
\[\mathcal L_{\rm f}(\psi)=\frac{1}{K}
  \sum_{i=1}^KS(\pi(\vartheta_i),\widehat\rho_{t}(\vartheta_i)),~where~
  \widehat\rho_{t}(\vartheta_i)=
  \frac{1}{K}
  \sum_{k=1}^K
  \sigma\!\left(
    \frac{R(\vartheta_i)-R(\widetilde\Theta_k)}{\tau}
  \right).\]

  \quad \quad   \textbf{Step 4.}  Aggregate the loss $\mathcal{L}(\psi) = \gamma_h\mathcal{L}_h(\psi)+\mathcal{L}_{d} (\psi)+\gamma_f\mathcal L_{\rm f}(\psi).$

\quad \quad  \textbf{Step 5.} Update $T_\psi$ based on $\nabla_{\psi} \mathcal L(\psi)$.

\noindent\colorbox[gray]{0.9}{\parbox{\dimexpr\linewidth-2\fboxsep\relax}{\centering\textbf{Inference}}}
Set $\widehat Q^{\mathrm{MIC}}:=Q_{\widehat\psi}$ after training. Any of the followings can be conducted individually depending on the context:

\quad \textbf{1. Bayesian-like Sampling} $\widetilde\Theta_1,...,\widetilde\Theta_M\sim \widehat Q^{\mathrm{MIC}}$ by $\widetilde\Theta_j=T_{\widehat\psi}(U_j)$ for $U_j\sim F_U.$

\quad \textbf{2. $100(1-\alpha)\%$ Credible Sampling} $\widehat C_\alpha = \{ T_{\widehat\psi}(U_j),~U_j\sim F_{U}(\,\cdot\mid\|U\|\leq 1-\alpha)\}.$

\quad \textbf{3. Contour Approximation} 
$\widehat\rho_M(\theta) = \frac{1}{M}\sum_{j=1}^M \mathbb{I}\bigl(R(\widetilde\Theta_j)\leq R(\theta)\bigr).$

\end{algorithm}

\begin{algorithm}
\caption{Amortized MIC Sampler Learning}
\label{alg:mic_amortized}
\spacingset{1.2}
\textbf{Input:} data $z$, rank function $R$.

\textbf{Tuning parameters:} loss weights $\gamma_{h},\gamma_f$, and temperature $(\tau)$.

\textbf{Learnable objects:} transport map $T_\psi$.

\textbf{Data:} $ D_{\rm target}=\{(z_i, \{\vartheta_{ib}, \pi_{z_i}(\vartheta_{ib})\}_{b=1}^N)\}_{i=1}^{N}$, $\vartheta_i\sim \operatorname{Unif}(\mathbb{T})$ computed for rank function $R$.

\noindent\colorbox[gray]{0.9}{\parbox{\dimexpr\linewidth-2\fboxsep\relax}{\centering\textbf{Transport Map Learning}}}

\textbf{for $t =1,...,N_{\mathrm{iter}}$ do}

 \quad \quad Sample { $(z_l, 
 \vartheta_{l1},\pi_{z_l}(\vartheta_{l1})),...,(z_l,\vartheta_{lB},\pi_{z_l}(\vartheta_{lB}))$ from $D_{\rm target}$ for $l=1,...,L$.}

 \quad \quad Sample $A_i\sim \operatorname{Unif}(0,1)$ and $V_j\sim \operatorname{Unif}(\mathbb{S}^{d-1})$.
 
\quad \quad  Define $U_{ij} = (1-A_i)V_j$ for $i\in[N]$ and $j\in[M]$.

\quad \quad  { Compute $\widetilde{\Theta}_{ijl}=T_\psi(U_{ij},z_l)$}.

\quad \quad  \textbf{Step 1.} Sample particles $U_1,\ldots,U_K\sim F_U$, and { transport by $\widetilde\Theta_{kl}=T_\psi(U_{k},z_l)$.}

\quad \quad Compute empirical loss $\mathcal{L}_{d} (\psi, z_l)= \frac{1}{MN}\sum_{i=1}^N \sum_{j=1}^M  S (A_i,\widehat\rho_{z_l}(\widetilde\Theta_{ijl}))$, where \[\widehat\rho_{z_l}(\widetilde\Theta_{ijl})=
  \frac{1}{K}
  \sum_{k=1}^K
  \sigma\!\left(
    \frac{R(\widetilde\Theta_{ijl},z_l)-R(\widetilde\Theta_{kl},z_l)}{\tau}
  \right).\]

  \quad \quad   \textbf{Step 2.} Compute a shell 1-nn distance by { $R_{ijl}^\psi = \min_{t\neq j} \| \widetilde{\Theta}_{ijl}-\widetilde{\Theta}_{itl}\|$ for each $(i,l)$}.

\quad \quad Compute empirical loss: 
\vspace{-0.5cm}
\[\mathcal{L}_{h}(\psi, z_l)=-\frac{d-1}{NM}\sum_{i=1}^N \sum_{j=1}^M \log R_{ijl}^\psi,\]

  \quad \quad     \textbf{Step 3.} Compute  $\mathcal L_{\rm f}(\psi,z_l)=\frac{1}{B}\sum_{b=1}^{B}S(\pi_{z_l}(\vartheta_{lb}),\widehat\rho_{z_l}(\vartheta_{lb}))$, where
  \[\widehat\rho_{z_l}(\vartheta_{lb})=
  \frac{1}{K}
  \sum_{k=1}^K
  \sigma\!\left(
    \frac{R(\vartheta_{lb},z_l)-R(\widetilde\Theta_{kl},z_l)}{\tau}
  \right).\]

  \quad \quad   \textbf{Step 4.}  Aggregate the loss $\mathcal{L}(\psi) = \frac{1}{L}\sum_{l=1}^L\left(\gamma_h\mathcal{L}_h(\psi,z_l)+\mathcal{L}_{d} (\psi,z_l)+\gamma_f\mathcal L_{\rm f}(\psi,z_l)\right).$

\quad \quad  \textbf{Step 5.} Update $T_\psi$ based on $\nabla_{\psi} \mathcal L(\psi)$.

\noindent\colorbox[gray]{0.9}{\parbox{\dimexpr\linewidth-2\fboxsep\relax}{\centering\textbf{Inference}}}
Set $\widehat Q^{\mathrm{MIC}}_{z}:=Q_{\widehat\psi,z}$ after training. Any of the followings can be conducted individually depending on the context. For the test point $z_0$,

\quad \textbf{1. Bayesian-like Sampling} $\widetilde\Theta_1,...,\widetilde\Theta_M\sim \widehat Q^{\mathrm{MIC}}$ by $\widetilde\Theta_j=T_{\widehat\psi}(U_j,z_0)$ for $U_j\sim F_U.$

\quad \textbf{2. $100(1-\alpha)\%$ Credible Sampling} $\widehat C_\alpha = \{ T_{\widehat\psi}(U_j,z_0),~U_j\sim F_{U}(\,\cdot\mid\|U\|\leq 1-\alpha)\}.$

\quad \textbf{3. Contour Approximation} 
$\widehat\rho_{z_0}(\theta) = \frac{1}{M}\sum_{j=1}^M \mathbb{I}\bigl(R(\widetilde\Theta_j,z_0)\leq R(\theta,z_0)\bigr).$

\end{algorithm}

\begin{algorithm}[t]
\caption{Data Generation for Amortized MIC Sampler Learning}
\label{alg:mic_amortized_data}
\spacingset{1.2}
\textbf{Input:} data $z_0$, likelihood function $L$, \# of parameter samples $N$, 
and MC step size $K$.

\textbf{Return:}
$D_{\rm target}
=
\left\{
\left(
Z_{\theta_i,k},
\theta_j,
\widehat{\pi}_{Z_{\theta_i,k}}(\theta_j)
\right)
:
i,j\in[N],~j\neq i,~k\in[K]
\right\}$.

\noindent
\colorbox[gray]{0.9}{
\parbox{\dimexpr\linewidth-2\fboxsep\relax}{
\centering\textbf{Data Generation}
}}
Define a parameter support (box) $\mathbb{T}$ based on $z_0$.

Sample
$\theta_1,\ldots,\theta_N
\sim \operatorname{Unif}(\mathbb{T})$.

For each $i\in[N]$, sample
$Z_{\theta_i,1},\ldots,Z_{\theta_i,K}
\overset{\mathrm{iid}}{\sim} P_{\theta_i}$,
and compute
$\widehat{\theta}^{\mathrm{MLE}}(Z_{\theta_i,k})$,
$k=1,\ldots,K$.

For each $i,k,$ define
$C_{i,k}
=
L\!\left(
\widehat{\theta}^{\mathrm{MLE}}(Z_{\theta_i,k}),
Z_{\theta_i,k}
\right)$
and
$R(\cdot,Z_{\theta_i,k})
=
L(\cdot,Z_{\theta_i,k})/C_{i,k}$. 

For each $j\in[N]$, define the empirical rank CDF
$\widehat F_j(r)
=
K^{-1}
\sum_{k'=1}^K
\mathbb{I}
\left\{
R(\theta_j,Z_{\theta_j,k'})
\leq r
\right\}$.

For each $Z_{\theta_i,k}$ and each $j\neq i$, compute
$\widehat{\pi}_{Z_{\theta_i,k}}(\theta_j)
=
\widehat F_j
\left\{
R(\theta_j,Z_{\theta_i,k})
\right\}$.

Return
$D_{\rm target}
=
\left\{
\left(
Z_{\theta_i,k},
\theta_j,
\widehat{\pi}_{Z_{\theta_i,k}}(\theta_j)
\right)
:
i,j\in[N],~j\neq i,~k\in[K]
\right\}$.
\end{algorithm}

\section{Experimental Details}

We train with Adam and a decaying learning rate initially set to $0.01$, using $\gamma_h=0.01$ and $\gamma_f=1$. We anneal $\tau$ to decrease to $0.05$. All neural networks use two hidden layers of width 64 with SiLU activations, denoted by MLP. This plain architecture was a deliberate choice so that the performance cannot be attributed to network design. For example, in Examples 1 and 3 (non-amortized versions), the network is not even given MLE information, so the network also needs to find the mode (MLE) during training. Consequently, it took quite a bit of time given the simplicity of the data. A well-designed network, for example, one centered at the MLE such as $T(u)=\hat\theta_z^{\rm MLE}+Wu+\|u\|\mathrm{MLP}(u,1-\|u\|)$, where $W$ is trainable, can make learning much faster, with 15-20 times fewer training steps. 

For amortized experiments (Example 2 and the ovarian application), to reduce the complexity of learning, we give the MLE information and use standardized parameter coordinates for shell entropy. Amortized samplers use $T(u, z) = \hat\theta_z^{\rm MLE} + \mathrm{MLP}\big([\,u,\; 1 - \|u\|,\; e(z)\,]\big)$, where $e(z)$ is a model-specific encoder of the data, and the output is applied in logit coordinates when the parameter box is constrained. 

For computation time, the S-shear model is a good example because its MLE is available in closed form. Once the training data is given, MIC training on a single RTX 3060 GPU takes 8.2 s per 1,000 steps for the non-amortized sampler and 18.5 s for the amortized sampler. Drawing 1,000 samples takes 0.29 ms and 0.16 ms, respectively, on the GPU. IMMC fitting takes 0.09 s per dataset, and drawing 1,000 samples takes 0.39 ms, both on the CPU. That is, once training or fitting is complete, the sampling times for MIC and IMMC are comparable. For models requiring expensive numerical MLE computation, a major computational cost for both methods is preparing the likelihood-ratio values needed for training or fitting.

\subsection{Example 1}\label{sec:details_example1}

We use the same Gamma and censored-Weibull data as
\citet{martin2025efficient}, which have 20 observations for Gamma and 26 observations including
14 censored times ($\Delta_i=1$) for Weibull. Weibull simulations use the estimated
censoring distribution obtained by Kaplan--Meier estimation.
For S-shear, we set $n=10$, $\sigma=2.5$, and the observed data are simulated at
$\theta=(0,0)$, with $\bar z=(-0.016,0.361)$.
The training and evaluation boxes are
$[1.002,3.197]\times[1.506,3.771]$ for Gamma,
$[-2,2]\times[6,10]$ for Weibull, and
$[-2.6993,2.6993]\times[-4.6993,4.6993]$ for S-shear.

MIC is trained for 300,000 optimization steps, using
1,000 uniformly sampled parameter targets per model. Training contour
values use 50 Monte Carlo datasets per target for Gamma and Weibull,
with likelihood maxima approximated over the target parameter set. IMMC uses the likelihood rank with
100 $\alpha$ levels and 500 Monte Carlo draws per level. For S-shear, both methods use the exact contour for training dataset. For all three datasets, both methods are fitted to the same observed data.

Evaluation uses a separate $50\times50$ grid over each box, with target
contours computed using 500 Monte Carlo datasets per point and MLE
normalization for Gamma and Weibull, and exactly for S-shear.
Each reconstructed contour $\hat\rho_Q$ uses $N_Q=10{,}000$ sampler draws.
MSE and maximum absolute error use the full grid. On the other hand, Spearman correlation
uses points with $\pi>0.005$.
The diagnostics $D_+$ and $D_-$ in Remark~\ref{rm:remark_Dplus} use
200 sampler draws, giving a 5\% two-sided DKW (Dvoretzky-Kiefer-Wolfowitz) reference threshold of
$\sqrt{\log(2/0.05)/(2\cdot200)}\approx0.096$ \citep{massart1990tight}.
The $C_{.05}$ coverage is the fraction of grid cells with $\pi\ge0.05$
occupied by at least one of the $N_Q$ draws. Draws are assigned to their
nearest grid point, where those outside the box are excluded.

\subsection{Example 2}\label{sec:exp2_details}

We use the same S-shear model and training box as in Example~1.
The training bank contains 100 datasets simulated at each of 1,000
uniformly sampled parameters, totaling 100,000 datasets.
The phase encoding used is $e(z) =(\sin(2.5 \bar z_1),\cos(2.5\bar z_1))$. We trained for 3,000 steps. The trained transport is fixed for all evaluations, whereas IMMC is
fitted separately to each dataset. Evaluation datasets are independent
of the training bank and shared across methods. Both fitted contours
use 5,000 reference draws per dataset; the reference contour is exact.
The three cases in Figure~\ref{fig:shear_amortize} have sample means
$(-0.751,-0.543)$, $(-0.377,0.371)$, and $(1.126,-0.668)$.
Their grid errors use a $140\times140$ grid. Estimating $D_\pm$ uses exact contour values at
12,000 sampler draws.

Coverage uses 200 datasets per phase, and power uses 100 per alternative.
Table~\ref{tab:amortized_cost} reports pointwise 95\% Wilson intervals
for coverage and medians with 10th/90th percentiles for region area.
Areas are computed analytically: if $\ell_\alpha$ is the
$(\lfloor\alpha M\rfloor+1)$-th smallest relative log-likelihood among
the $M=5{,}000$ reference draws, the fitted region has area
$-2\pi\sigma^2\min(\ell_\alpha,0)/n$. The exact area is
$-2\pi\sigma^2\log(\alpha)/n$, since $m$ has unit Jacobian.

For the predictive check, we use the S-shear data from Example~1. The larger-variance data are generated as $\bar z+1.5(z_i-\bar z)$. For the mixture data, we draw five observations from each of two Gaussian components with means $\pm(2.29,2.29)$ and covariance $I_2$, then recenter the pooled sample to $\bar z$. The equally weighted mixture has marginal variances $2.29^2+1\approx\sigma^2$. All three datasets share the same sample mean and hence the same exact contour.  

\subsection{Example 3}\label{sec:exp3_details}

We use the third dataset of Example~2, where the fixed data sample $z$ was such as $\bar z=(1.126,-0.668)$.
MIC is trained the same as in Example 1. IMMC uses the settings of Example~1 on the same data. The Reference uses the exact contour and uniform Euclidean arclength
on each level curve in $\theta$ coordinates.
Numerical integration uses 200 equal depth bins, three quadrature
levels per bin, and 2,048 arclength-equispaced points per level.
The fraction of points in $H$ gives the spectrum, from which $Q(H)$
and mean depth are integrated.  For reference belief and plausibility, we compute the supremum by direct numerical maximization since the exact contour is available in closed form.

MIC and IMMC each use 200,000 sampler draws and an independent set of
5,000 reference draws for the stitched contour.
$Q(H)$ and mean depth are estimated by the sample fraction in $H$ and
the average stitched depth among those samples; the spectrum uses
200 depth bins. Belief and plausibility are computed from stitched-depth
maxima over $H^c$ and $H$, using a common pool consisting of both
methods' sampler draws, a $200\times200$ grid over the box, and the MLE.
The selection rule is applied to all 15 candidate unions with
$\alpha_0=0.10$ and $\beta=0.30$.

\subsection{Ovarian Aging Analysis}
\paragraph{Model.} The detection limit of the AMH assay used here is $d = 0.0018$ ng/mL \cite{de2016back}, and values below it are recorded only as ``below $d$.'' What we actually observe is $\delta_{ij} = \mathbf{1}\{Y^*_{ij} < \log d\}, $ $Y_{ij} = \max(Y^*_{ij}, \log d)$
and a censored measurement enters the likelihood through a probability rather than a density. Writing $\mu_{ij}(c,a) = f(t_{ij}+c) + a$, the likelihood for woman $i$ is
\begin{equation}
L_i(c,a) = \prod_{j:\,\delta_{ij}=0} \frac{1}{\sigma}\,\phi\!\left(\frac{Y_{ij}-\mu_{ij}(c,a)}{\sigma}\right)
\prod_{j:\,\delta_{ij}=1} \Phi\!\left(\frac{\log d-\mu_{ij}(c,a)}{\sigma}\right)
\end{equation}
A censored observation tells us only that ``this person's latent measurement was below the detection limit at this age.'' When $f$ is decreasing, with $a$ fixed, this observation's likelihood contribution increases as $c$ increases (as the ovarian clock moves further ahead) and decreases as $c$ decreases. Once the latent mean is sufficiently below the detection limit, this contribution approaches 1 and does not distinguish well how much further below the limit it is.
\paragraph{Rationale for $b = 0.946$ ng/mL.}
This is almost the same as the value of the population curve at age 40 ($e^{f(40)} \approx 0.96$ ng/mL), and is a familiar value because it falls within the AMH cut-off range (0.5--1.1 ng/mL) of the Bologna criteria widely used to predict poor response in IVF \citep{ferraretti2011eshre}. This is an analytical reference value, however, rather than a clinical decision threshold, and is unrelated to the chance of natural conception. We consider $k$ from 0 to 30 years in one-year increments.

\begin{figure}
    \centering
    \includegraphics[width=0.7\linewidth]{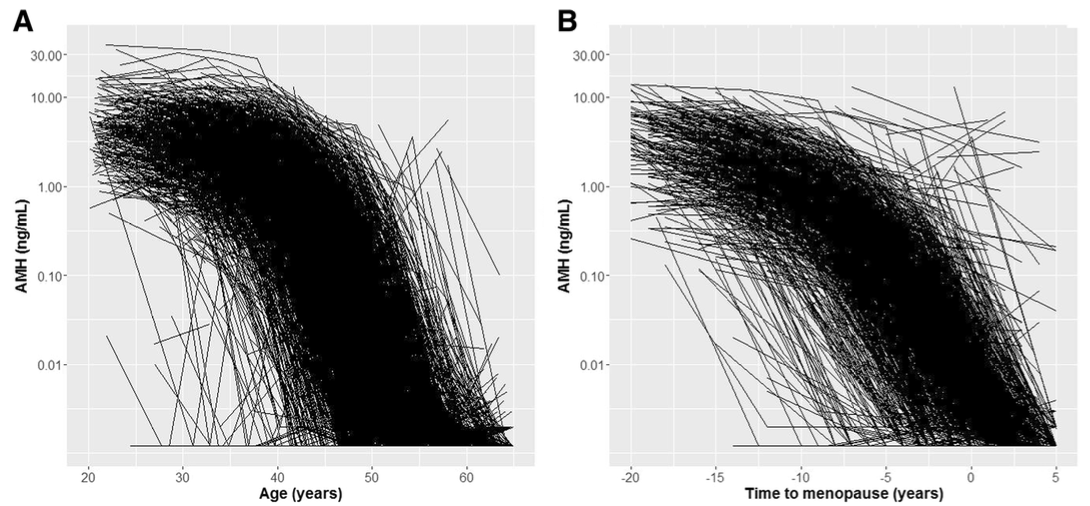}
    \includegraphics[width=0.7\linewidth]{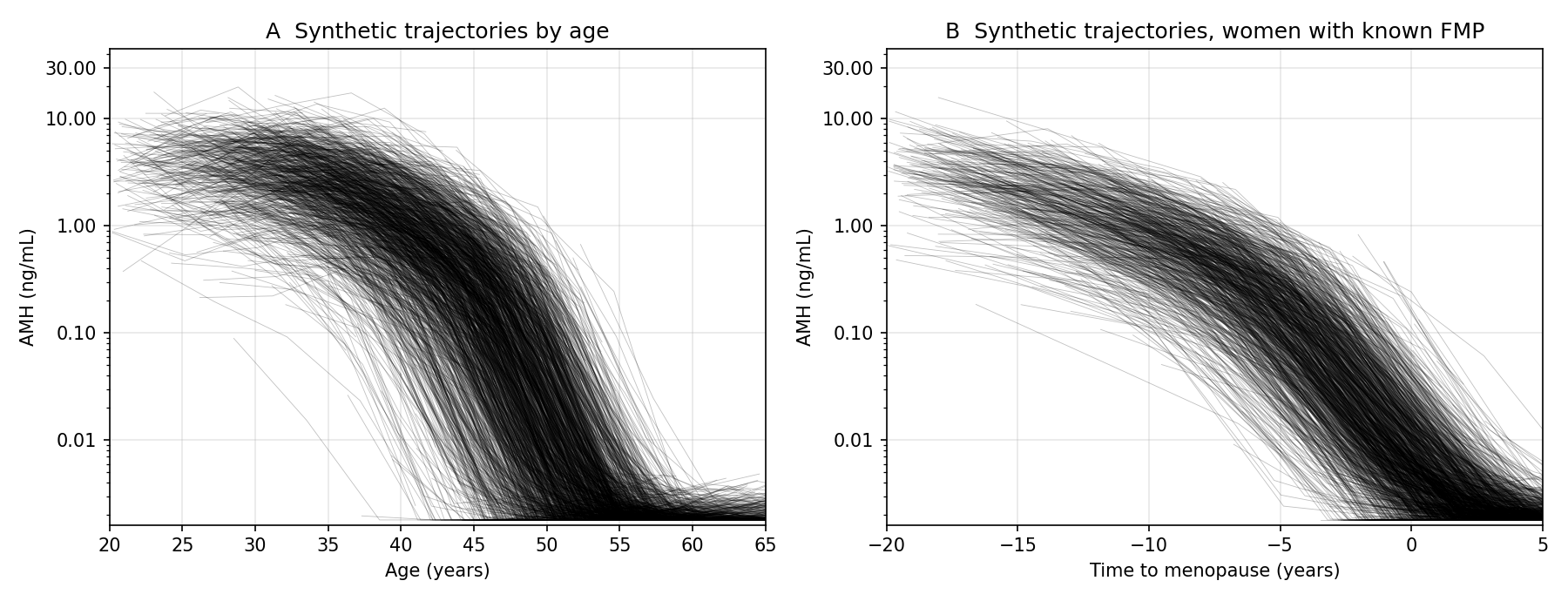}
    \caption{The comparison between the original and the synthetic trajectory. Top: The original, copied from \cite{de2016back}. Bottom: From our synthesized data.}
    \label{fig:compare_digital}
\end{figure}

\paragraph{Data.}
Since individual records from \cite{de2016back} are not publicly available, we constructed synthetic longitudinal AMH records using their reported age-specific quantile curves and summary statistics. We represented the population reference curve $f$ by shape-preserving interpolation and drew $c_i$ and $a_i$ independently from mean-zero skew-normal distributions. The curve and distribution parameters were calibrated jointly to broadly reproduce the reported age-specific quantiles and proportions of measurements below the detection limit. Measurements were then generated from \eqref{eq:amh_model} and censored at the detection limit $d$. We followed the cohort design in \cite{de2016back}, matching the reported initial age-group proportions, five-year visit spacing, and participant counts across successive rounds. Of the 3,326 synthetic women, 371 had at least four measurements between ages 20 and 45; we used their first four measurements ($n=4$). Figure~\ref{fig:compare_digital} compares the age-specific AMH distribution in the full synthetic cohort with the published curves.  In our analysis, we focus on the 160 women out of the 371, whose MLEs imply a positive remaining time.

\paragraph{Computational details.}
The encoder $e(z)$ is simply the MLE (rescaled to $[-1,1]^2$) concatenated with the data $z$ itself; the first-visit age $(t_{i1} - 25)/5$, the four log-AMH measurements standardized by the mean and standard deviation of the training bank (censored values enter as $\log d$), and the four censoring indicators $\delta_{ij}$. IMMC also receives these variables. The training bank contains 1,000 parameters sampled uniformly from $[-15,15]\times[-4,2]$ and 200 datasets per parameter. We train for 40,000 steps. IMMC is fitted separately to each woman. Both methods use 10,000 draws for stitched contours. The MC reference uses 500 simulated datasets at each point of a $151\times61$ grid over the parameter box, preserving each woman's visit ages and applying the censoring rule. Relative-likelihood ranks use a numerically computed box-constrained MLE for each simulated dataset. The reference contour is bilinearly interpolated onto a $301\times121$ grid; When taking supremum to compute Bel and Pl, the candidate theta pool includes the MLE and points immediately on either side of each hypothesis boundary besides the grid.

\section{Additional Figures and Tables}
This section provides additional supportive results.
\begin{figure}
    \centering
    \includegraphics[width=1\linewidth]{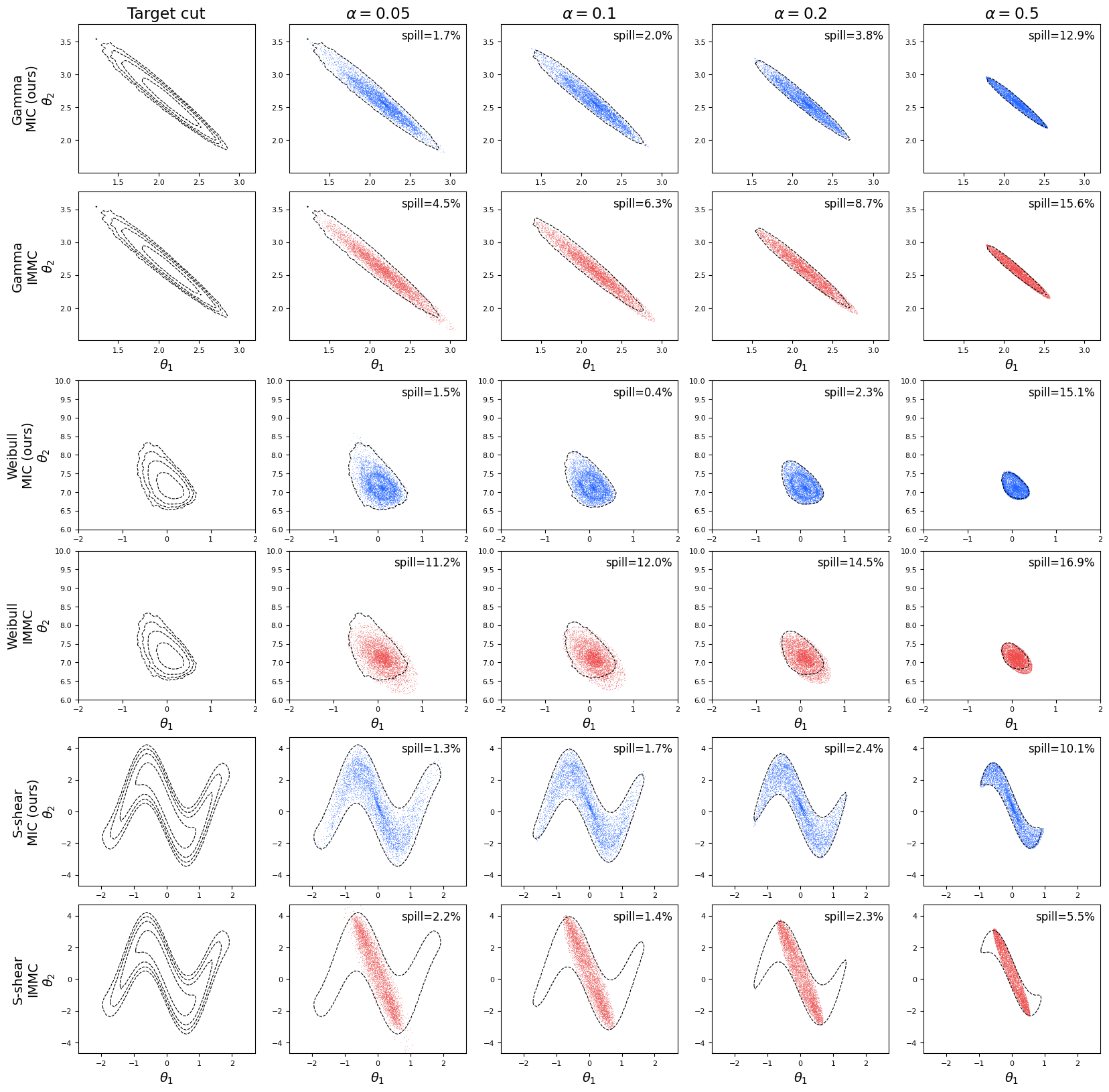}
    \caption{(Example~1) Samples from MIC (blue) and IMMC (red) generated with the various source level $\alpha$. The black dotted lines are estimated target $C_\alpha$. }
    \label{fig:full_nested_contour}
\end{figure}

\begin{figure}
    \centering
    \includegraphics[width=0.9\linewidth]{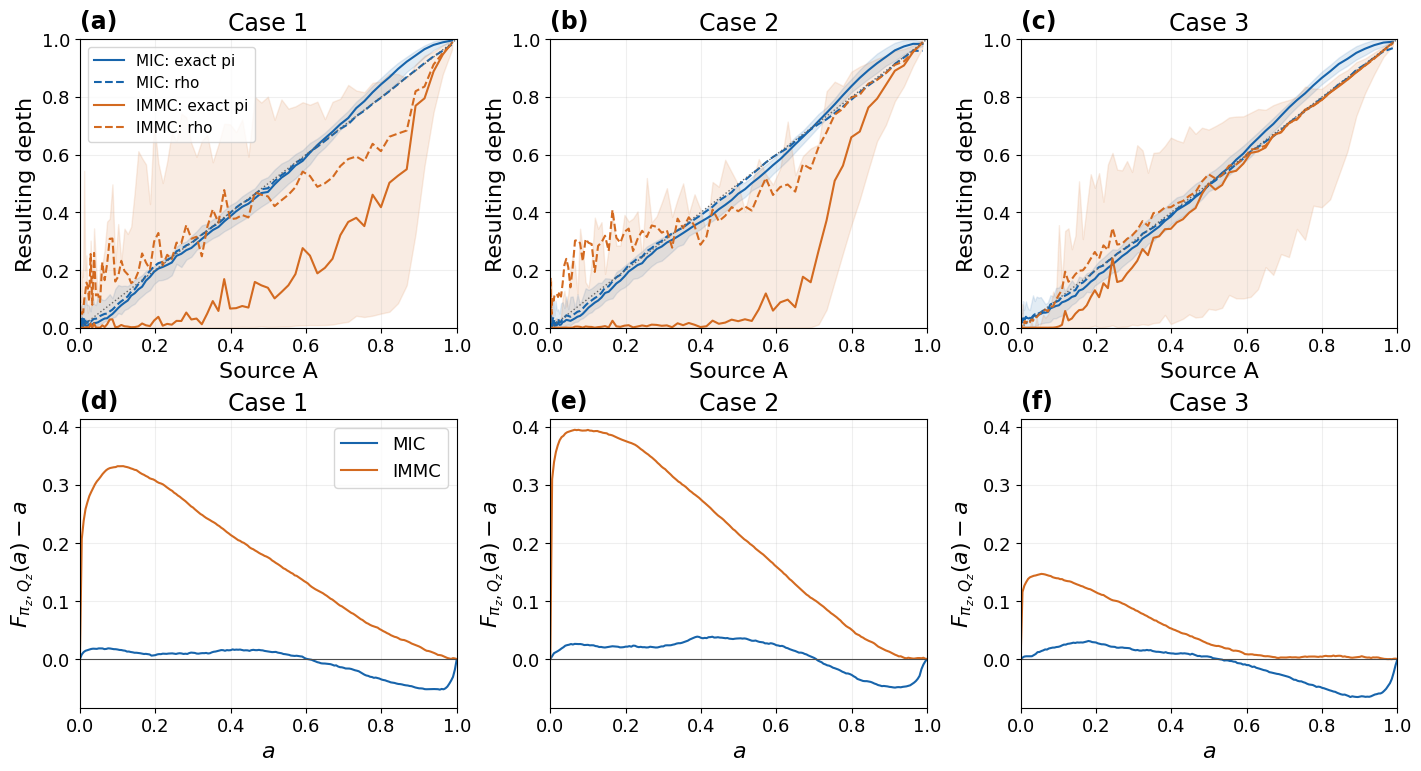}
    \caption{(Example 2) The source radius to contour depth alignment (top) and the credal calibration (bottom). The contour depth is computed based on the known true contour and the stitched estimation.}
    \label{fig:sub1}
\end{figure}

\begin{table}[h!]
\centering
\small
\caption{(Example 3) Selected unions among all 15 candidates, with $\widehat{\mathrm{Bel}}(H)\geq 0.90$ and $\widehat Q(H)-\widehat{\mathrm{Bel}}(H)\leq 0.30$. Gap denotes $\widehat Q(H)-\widehat{\mathrm{Bel}}(H)$, ${\dagger}$ indicates infeasibility, and $\star$ marks the selected hypothesis. MIC and Reference selected $H_4$ and IMMC selected $H_{1,2,3,4}$.}
\label{tab:belpl-winners}
\begin{tabular}{llrrrrr}
\toprule
Method & $H_S$ & Bel & Pl & $Q(H)$ & Gap & $\bar\pi$ \\
\midrule
Reference & $H_{4}$ & 0.93172 & 1.00000 & 0.98540 & 0.05368 & $\mathbf{0.50702}^{\star}$ \\
MIC & $H_{4}$ & 0.93140 & 1.00000 & 0.95814 & 0.02674 & $\mathbf{0.51230}^{\star}$ \\
IMMC & $H_{4}$ & 0.79780 & 1.00000 & 0.83044 & 0.03264 & 0.58206$^{\dagger}$ \\
\midrule
Reference & $H_{1,2,3,4}$ & 0.99873 & 1.00000 & 0.99998 & 0.00125 & 0.50001 \\
MIC & $H_{1,2,3,4}$ & 0.99980 & 1.00000 & 1.00000 & 0.00020 & 0.49190 \\
IMMC & $H_{1,2,3,4}$ & 0.90160 & 1.00000 & 0.96544 & 0.06384 & $\mathbf{0.51622}^{\star}$ \\
\bottomrule
\end{tabular}
\end{table}

\begin{figure}
    \centering
    \includegraphics[width=01\linewidth]{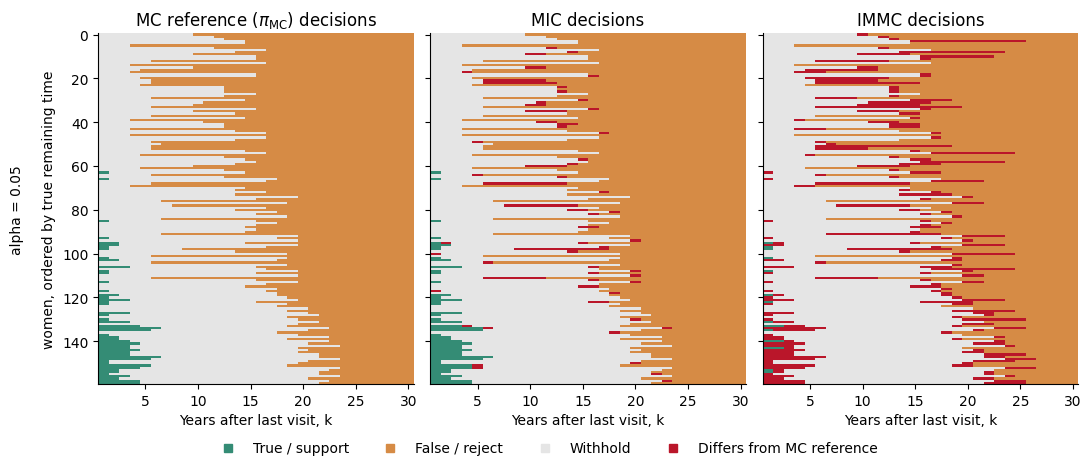}
    \caption{(Ovarian Application) Comparison to the MC reference decisions. MIC mostly aligns to the MC reference.}
    \label{fig:MC_ovarian}
\end{figure}

\begin{table}[htbp]
\centering
\small
\caption{(Ovarian Application) Hypothesis decisions at $\alpha=0.05$ for the 160 women of the primary cohort. An error is an incorrect acceptance or rejection of $H_k$. In the ``All horizons'' row, the withhold rate is averaged over the full grid $k=1,\ldots,30$, and the error count records women with at least one error anywhere on that grid. MC ref.\ is the direct Monte Carlo contour on the parameter grid.}
\label{tab:amh_decision_details}
\begin{tabular}{lrrrrrr}
\toprule
& \multicolumn{3}{c}{Withhold rate (\%)} & \multicolumn{3}{c}{Women with an error} \\
\cmidrule(lr){2-4}\cmidrule(lr){5-7}
Horizon (years) & MIC & IMMC & MC ref. & MIC & IMMC & MC ref. \\
\midrule
1 & 68.8 & 91.2 & 68.8 & 0 & 0 & 0 \\
3 & 86.9 & 100.0 & 86.9 & 0 & 0 & 0 \\
5 & 88.1 & 96.2 & 89.4 & 0 & 0 & 0 \\
10 & 83.1 & 88.1 & 76.2 & 0 & 0 & 0 \\
\midrule
All horizons & 46.0 & 54.4 & 43.9 & 0 & 0 & 0 \\
\bottomrule
\end{tabular}
\end{table}

\begin{figure}
    \centering
    \includegraphics[width=1\linewidth]{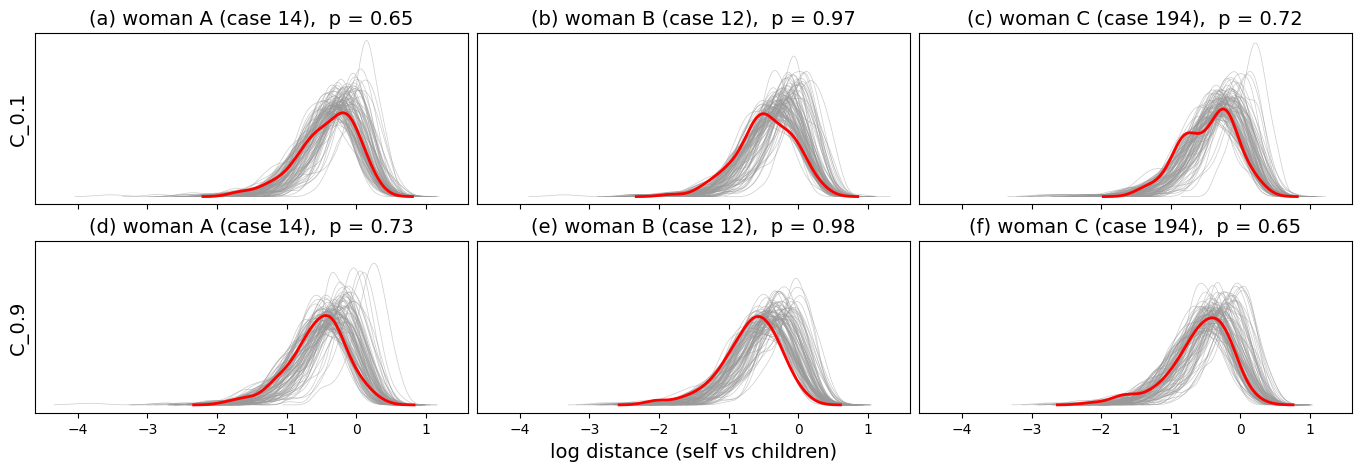}
    \caption{(Ovarian Application) Predictive check result for three chosen (simulated) women. It gives an illustration of the predictive check when the data is well explained by the likelihood model and when the amortized sampler is well-trained. Their diagnostics value is far from 0 and the red line (self-children relationship) resembles the other gray lines (children-grand-children relationship).}
    \label{fig:ovarian_predictive}
\end{figure}

\begin{figure}
    \centering
    \includegraphics[width=1\linewidth]{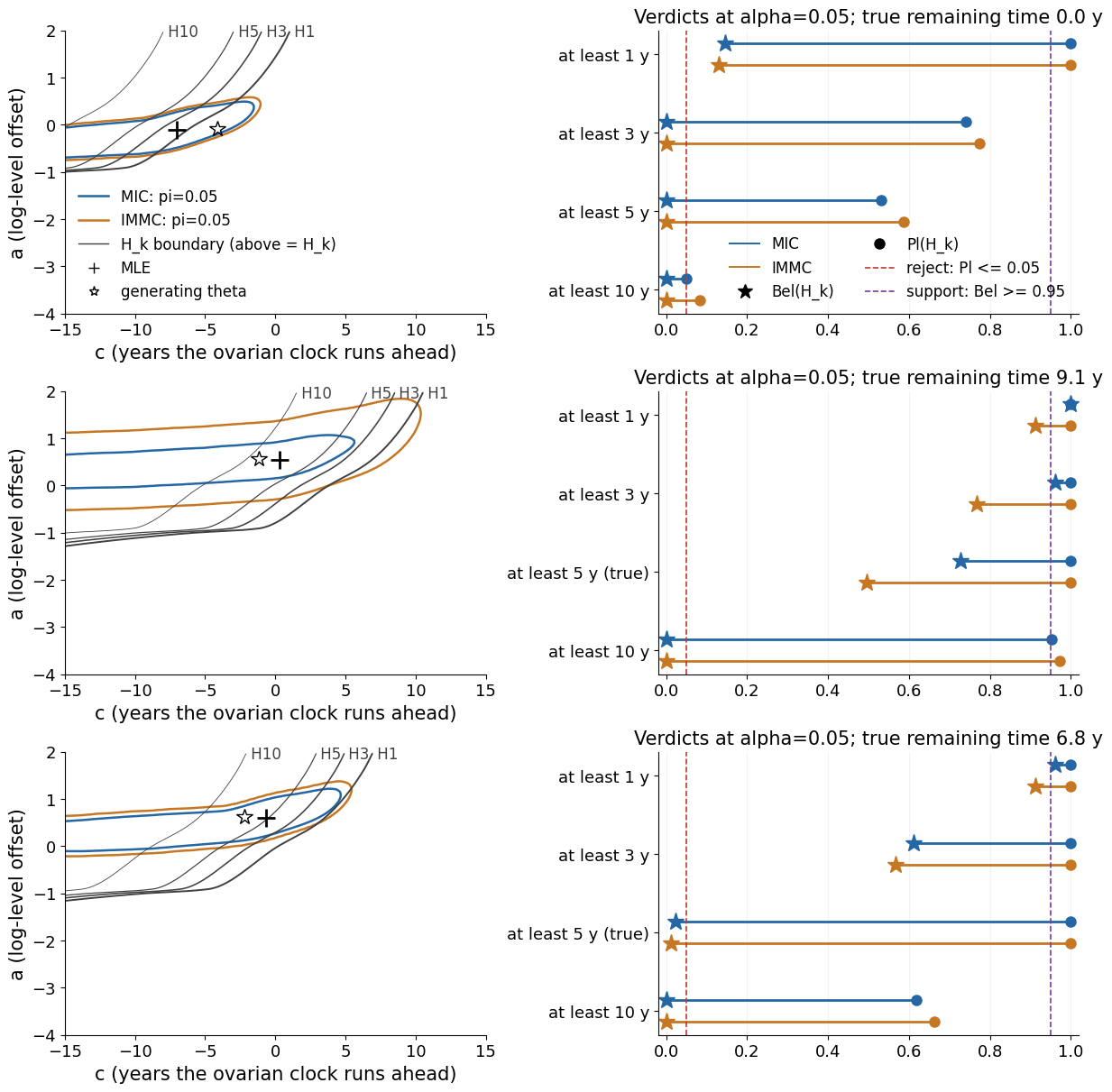}
    \caption{(Ovarian Application) A few chosen individual examples. Left: Multiple hypotheses and estimated contours. Right: Bel and Pl values showing acceptance/rejection with $\alpha=0.05$. }
    \label{fig:amh_spectrum}
\end{figure}

\end{document}